\documentclass{article}

\usepackage[final, nonatbib]{neurips_2021}
\usepackage[numbers]{natbib}

\usepackage[utf8]{inputenc} 
\usepackage[T1]{fontenc}    
\usepackage{hyperref}       
\usepackage{url}            
\usepackage{booktabs}       
\usepackage{amsfonts}       
\usepackage{nicefrac}       
\usepackage{microtype}      
\usepackage{xcolor}         
\usepackage{adjustbox}

\usepackage{amsmath,amsthm,amssymb}
\usepackage{algorithm}
\usepackage{algorithmic}
\usepackage{multicol}
\usepackage{comment}
\usepackage{mathtools}
\usepackage{caption}
\usepackage{float}
\usepackage{graphicx}
\usepackage{subfigure}
\usepackage{multirow}
\usepackage{mathrsfs}
\usepackage{newfloat}
\usepackage{relsize}
\usepackage{enumitem}
\usepackage{nicefrac}
\usepackage{pifont}
\usepackage{wrapfig}
\usepackage{bm}

\newcommand{\cmark}{\color{green}\ding{51}}%
\newcommand{\xmark}{\color{red}\ding{55}}%

\newcommand{\D}{\mathcal{D}}
\newcommand{\Dtr}{\mathcal{D}_{\text{train}}}
\newcommand{\Dte}{\mathcal{D}_{\text{test}}}
\newcommand{\Emax}{E_{\text{max}}}

\newcommand{\R}{\mathbb{R}}

\newcommand{\E}{\mathbb{E}}

\newcommand{\colin}[1]{\textcolor{red}{[#1]}}
\renewcommand\vec{\bm}

\title{NAS-Bench-x11 and the Power of Learning Curves}

\author{Shen Yan\thanks{
Equal contribution. Correspondence to
\texttt{yanshen6@msu.edu}, \texttt{colin@abacus.ai}, \texttt{ysavani@cs.cmu.edu}, \texttt{fh@cs.uni-freiburg.de}.
}~ $^1$, Colin White$^{*2}$, Yash Savani$^3$, Frank Hutter$^{4,5}$\\
    $^1$ Michigan State University, $^2$ Abacus.AI, 
    $^3$ Carnegie Mellon University, \\
    $^4$ University of Freiburg, 
    $^5$ Bosch Center for Artificial Intelligence
  }

\begin{document}

\maketitle

\begin{abstract}
While early research in neural architecture search (NAS) required extreme computational resources,
the recent releases of tabular and surrogate benchmarks
have greatly increased the speed and reproducibility of NAS research. 
However, two of the most popular benchmarks do not provide the full training information for each architecture. 
As a result, on these benchmarks it is not possible to run many types of
multi-fidelity techniques, such as learning curve extrapolation, 
that require evaluating architectures at arbitrary epochs.
In this work, we present a method using singular value decomposition and noise modeling 
to create surrogate benchmarks, NAS-Bench-111, NAS-Bench-311, and NAS-Bench-NLP11, 
that output the full training information for each architecture,
rather than just the final validation accuracy.
We demonstrate the power of using the full training information by introducing a 
learning curve extrapolation framework to modify single-fidelity algorithms, showing that it leads 
to improvements over popular single-fidelity algorithms which claimed to be state-of-the-art upon release.
Our code and pretrained models are available at \url{https://github.com/automl/nas-bench-x11}.
\end{abstract}

\section{Introduction} \label{sec:intro}
In the past few years, algorithms for neural architecture search (NAS) have been used to 
automatically find architectures that achieve state-of-the-art performance on various datasets~\citep{zoph2017neural, real2019regularized, darts, nas-survey}.
In 2019, there were calls for reproducible and fair comparisons within NAS 
research~\citep{randomnas, sciuto2019evaluating,yang2019evaluation, lindauer2019best} due to both the lack of a consistent training pipeline between papers and experiments with not enough trials 
to reach statistically significant conclusions.
These concerns spurred the release of tabular benchmarks, such as NAS-Bench-101~\citep{nasbench}
and NAS-Bench-201~\citep{nasbench201}, created by fully training all architectures in search
spaces of size $423\,624$ and $6\,466$, respectively. 
These benchmarks allow researchers to easily
simulate NAS experiments, making it possible to run fair NAS comparisons and to run enough trials 
to reach statistical significance at very little computational cost or carbon 
emissions~\citep{hao2019training}.
Recently, to extend the benefits of tabular NAS benchmarks to larger, more realistic NAS search spaces which cannot be evaluated exhaustively, it was proposed to construct 
\emph{surrogate benchmarks}~\citep{nasbench301}. 
The first such surrogate benchmark is NAS-Bench-301~\citep{nasbench301}, which
models the DARTS~\citep{darts} search space of size $10^{18}$ architectures. It 
was created by fully training $60\,000$ architectures (both drawn randomly and
chosen by top NAS methods) 
and then fitting a surrogate model
that can estimate the performance of all of the remaining architectures.
Since 2019, dozens of papers have used these NAS benchmarks to develop new 
algorithms~\citep{bananas, nasbowl, yan2021cate, npenas, shi2020bonas}.

An unintended side-effect of the release of these benchmarks is that
it became significantly easier to devise \emph{single fidelity} NAS algorithms: 
when the NAS algorithm chooses to evaluate an architecture, the architecture is fully trained
and only the validation accuracy at the final epoch of training is outputted.
This is because NAS-Bench-301 only contains the architectures' accuracy at epoch 100,
and NAS-Bench-101 only contains the accuracies at epochs 4, 12, 36, and 108
(allowing single fidelity or very limited multi-fidelity approaches).
NAS-Bench-201 does allow queries on the entire learning curve (every epoch),
but it is smaller in size ($6\,466$) than NAS-Bench-101 
($423\,624$) or NAS-Bench-301 ($10^{18}$).
In a real world experiment, since training architectures to convergence is computationally
intensive, researchers will often run \emph{multi-fidelity} algorithms: 
the NAS algorithm can train architectures to any desired epoch.
Here, the algorithm can make use of speedup techniques such as learning curve 
extrapolation (LCE)~\citep{swersky2014freeze,domhan2015speeding,baker2017accelerating, lcnet}
and successive halving~\citep{hyperband,bohb,li2018massively,abohb}.
Although multi-fidelity techniques are often used in the hyperparameter optimization
community~\citep{k2016multifidelity, k2017multifidelity, hyperband, bohb, swersky2014freeze},
they have been under-utilized by the NAS community in the last few years.

\begin{figure}[t]
	\centering
    \includegraphics[width=.24\columnwidth]{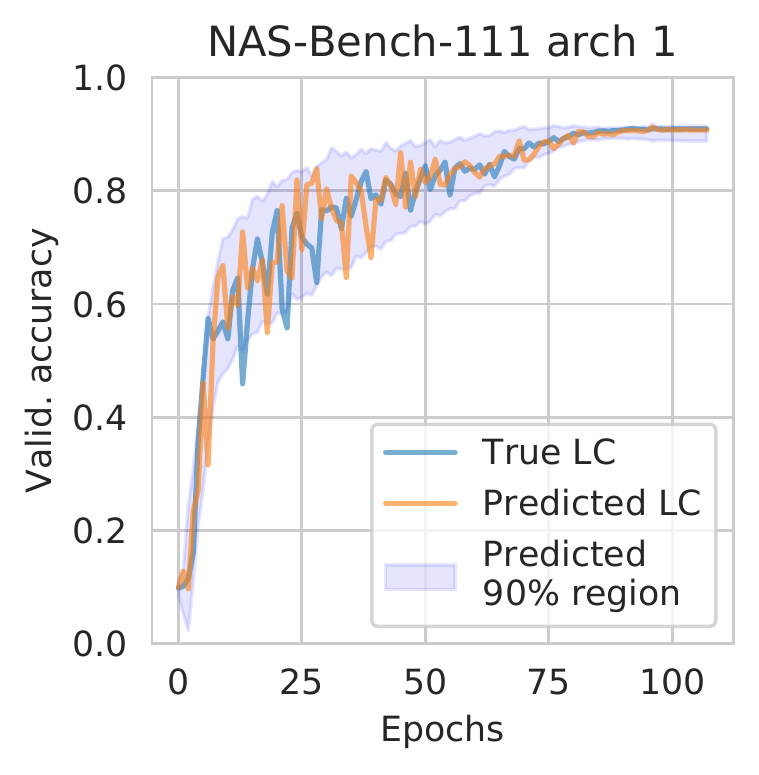}
    \includegraphics[width=.24\columnwidth]{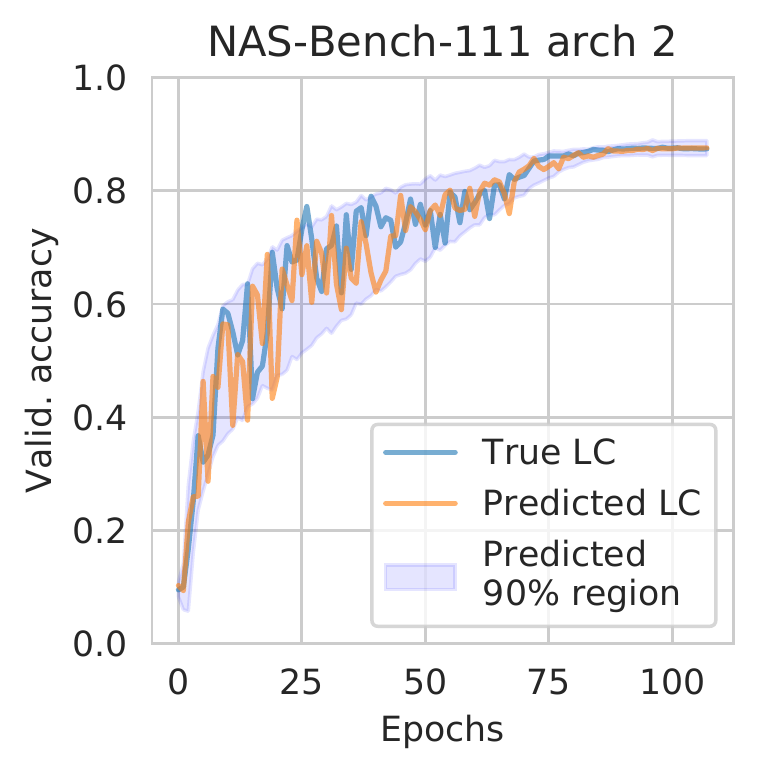}
    \includegraphics[width=.24\columnwidth]{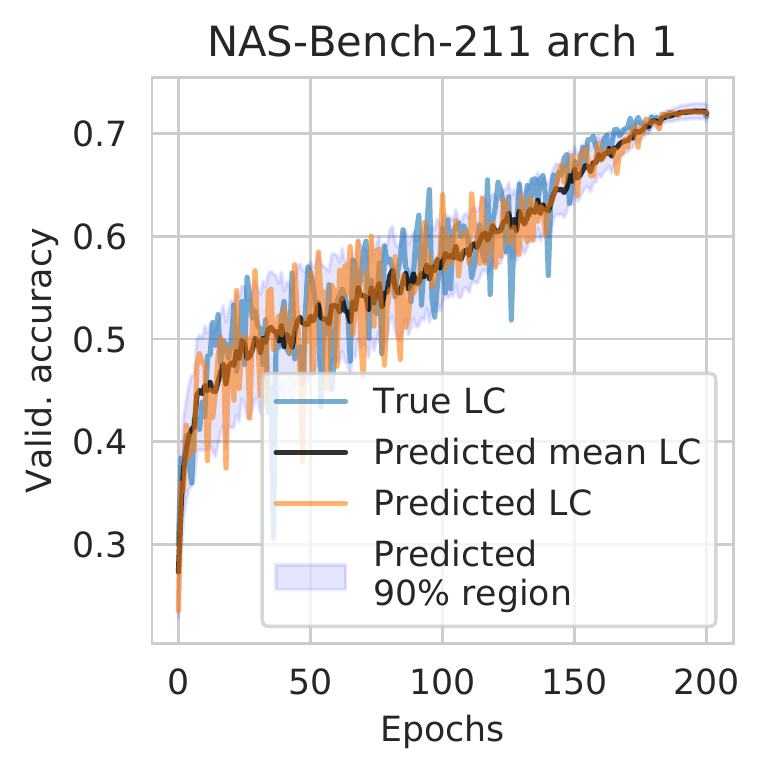}
    \includegraphics[width=.24\columnwidth]{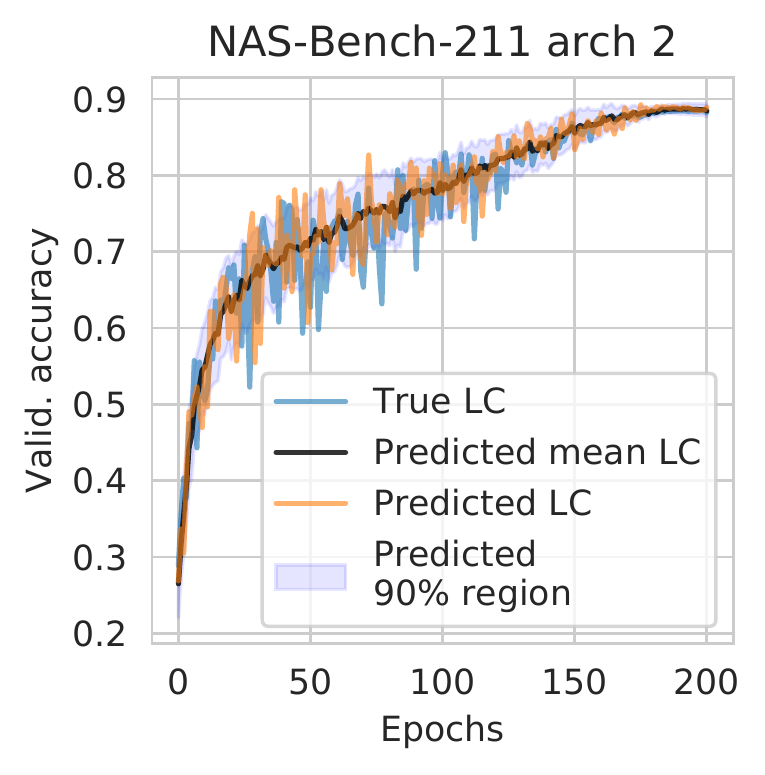}
    \includegraphics[width=.24\columnwidth]{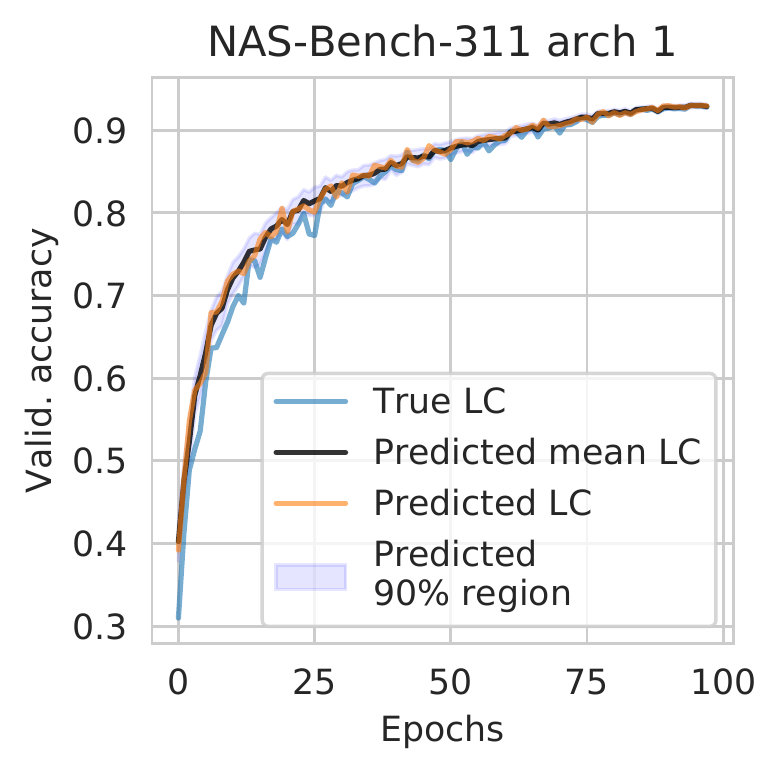}
    \includegraphics[width=.24\columnwidth]{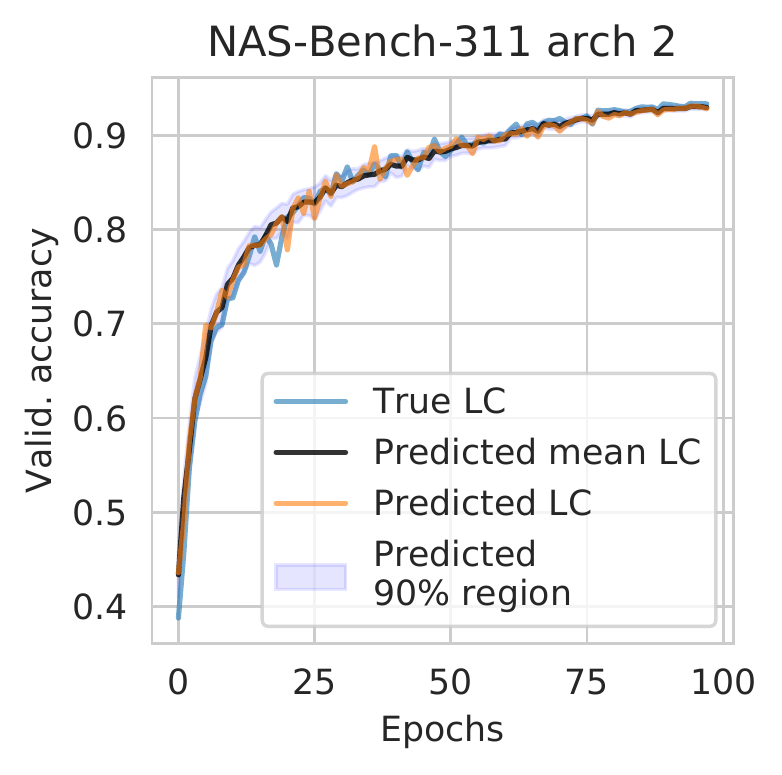} 
    \includegraphics[width=.24\columnwidth]{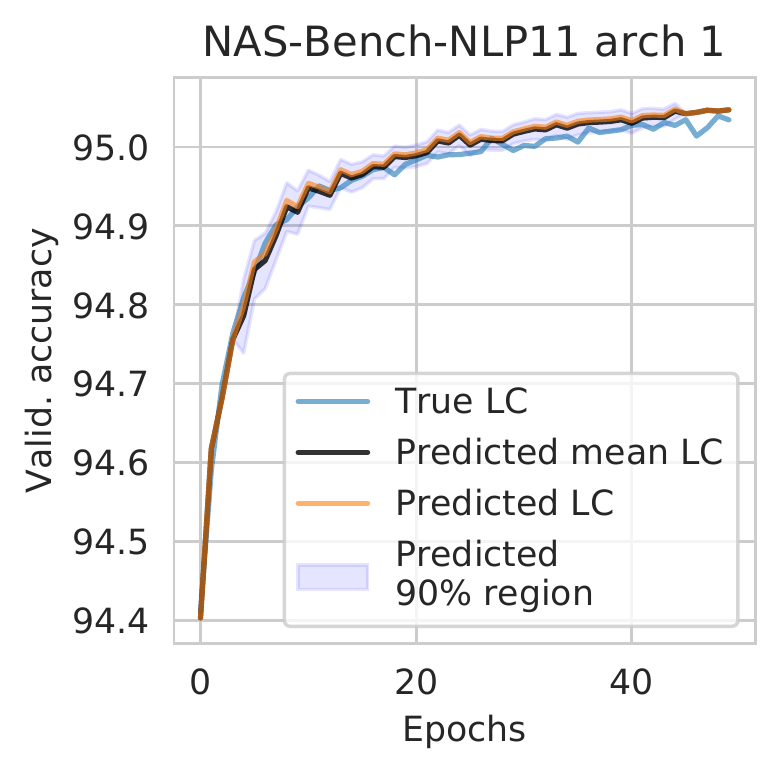}
    \includegraphics[width=.24\columnwidth]{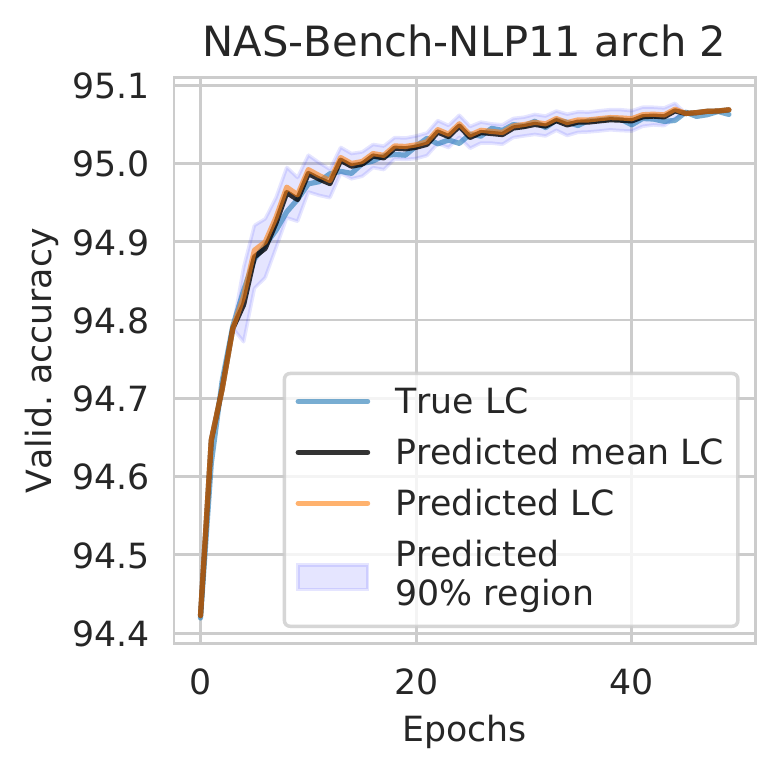}
    \caption{Each image shows the true learning curve vs.\ the learning curve predicted by
    our surrogate models, with and without predicted noise modeling. We also
    plot the 90\% confidence interval of the predicted noise distribution.
    We plot two architectures each for NAS-Bench-111, NAS-Bench-211, NAS-Bench-311, and NAS-Bench-NLP11.}
	\label{fig:lcs}
\end{figure}

In this work, we fill in this gap by releasing NAS-Bench-111, NAS-Bench-311, and NAS-Bench-NLP11,
surrogate benchmarks with full learning curve information for train, validation, 
and test loss and accuracy for all architectures, significantly extending NAS-Bench-101, 
NAS-Bench-301, and NAS-Bench-NLP~\citep{nasbenchnlp}, respectively.
With these benchmarks, researchers can easily incorporate multi-fidelity techniques, such
as early stopping and LCE into their NAS algorithms. 
Our technique for creating these benchmarks can be summarized as follows.
We use a training dataset of architectures (drawn randomly and chosen by top NAS methods) with good coverage over the search space, along with full learning curves, to fit a model that
predicts the full learning curves of the remaining architectures. We employ three techniques to fit 
the model: \emph{(1)}  dimensionality reduction of the learning curves, 
\emph{(2)} prediction of the top singular value coefficients, 
and \emph{(3)} noise modeling. These techniques can be used in the future to create new NAS 
benchmarks as well.
To ensure that our surrogate benchmarks are highly accurate, 
we report statistics such as
Kendall Tau rank correlation and Kullback Leibler divergence between ground truth learning curves
and predicted learning curves on separate test sets.
See Figure~\ref{fig:lcs} for examples of predicted learning curves on the test sets.

To demonstrate the power of using the full learning curve information, 
we present a framework 
for converting single-fidelity NAS algorithms into multi-fidelity algorithms using
LCE. We apply our framework to popular single-fidelity NAS algorithms, such as regularized 
evolution~\citep{real2019regularized}, local search~\citep{white2020local}, and 
BANANAS~\citep{bananas}, all of which claimed state-of-the-art upon release, showing that they can be further improved across four search spaces. Finally, we also benchmark multi-fidelity algorithms such as Hyperband~\citep{hyperband} and BOHB~\citep{bohb} alongside single-fidelity algorithms. 
Overall, our work bridges the gap between different areas of AutoML
and will allow researchers to easily develop effective multi-fidelity and
LCE techniques in the future.
To promote reproducibility, we release our code and 
we follow the NAS best practices checklist~\citep{lindauer2019best}, providing the details in Appendix~\ref{app:nas_checklist}.

\noindent\textbf{Our contributions.}
We summarize our main contributions below.
\begin{itemize}[topsep=0pt, itemsep=2pt, parsep=0pt, leftmargin=5mm]
    \item 
    We develop a technique to create surrogate NAS benchmarks that include the
    full training information for each architecture, including train, validation, and
    test loss and accuracy learning curves. This technique can be used to create future NAS benchmarks on any search space.
    \item
    We apply our technique to create NAS-Bench-111, NAS-Bench-311, and NAS-Bench-NLP11,
    which allow researchers to easily develop multi-fidelity NAS algorithms that achieve higher performance than single-fidelity techniques.
    \item 
    We present a framework for converting single-fidelity NAS algorithms into 
    multi-fidelity NAS algorithms using learning curve extrapolation, and we show that our
    framework allows popular state-of-the-art NAS algorithms to achieve further improvements.
\end{itemize}

\section{Related Work} \label{sec:relatedwork}
NAS has been studied since at least the late
1980s~\citep{miller1989designing, kitano1990designing, stanley2002evolving} 
and has recently seen a 
resurgence~\citep{zoph2017neural,negrinho2017deeparchitect,enas,nasbot,real2019regularized,hu2019forwardnas}.
Weight sharing algorithms have become popular due to their computational 
efficiency~\citep{bender2018understanding, darts, gdas, zela2020understanding, you2020greedynas, peng2020cream, yu2021landmark}.
Recent advances in performance 
prediction~\citep{wen2019neural, ning2020surgery, shi2020bonas, yan2020does, seminas, white2021powerful, nasbowl, bananas} and other iterative techniques~\citep{bohb,vu2020bayesian} have reduced the runtime gap between iterative and weight sharing techniques. 
For detailed surveys on NAS, we suggest referring to~\citep{nas-survey,xie2020weight}. 

\paragraph{Learning curve extrapolation.}
Several methods have been proposed to estimate the final validation accuracy
of a neural network by extrapolating the learning curve of a partially trained
neural network.
Techniques include fitting the partial curve to an ensemble of various
parametric
functions~\citep{domhan2015speeding}, predicting the performance based on the derivatives of the learning curves of partially trained neural network configurations~\citep{baker2017accelerating}, summing the training losses~\citep{ru2020revisiting},
using the basis functions as the output layer of a Bayesian neural network~\citep{lcnet}, using previous learning curves as basis function extrapolators~\citep{chandrashekaran2017speeding}, using the positive-definite covariance kernel to capture a variety of training curves~\citep{swersky2014freeze},  or using a Bayesian recurrent neural network~\citep{gargiani2019probabilistic}. While in this work we focus on multi-fidelity optimization utilizing learning curve-based extrapolation, another main category of methods lie in bandit-based algorithm selection~\citep{hyperband,bohb,abohb,huang2020asymptotically,dehb}, and the fidelities can be further adjusted according to the previous observations or a learning rate scheduler~\citep{k2016multifidelity,k2017multifidelity,klein17fast}.

\begin{table}[t]
\caption{Overview of existing NAS benchmarks. We introduce NAS-Bench-111, -311, and -NLP11. 
}
\vspace{1mm}
\centering
\begin{tabular}{@{}l|c|c|c|c@{}}
\toprule
\multicolumn{1}{l}{\textbf{Benchmark}} & \multicolumn{1}{c}{\textbf{Size}} & \multicolumn{1}{c}{\textbf{Queryable}} & \multicolumn{1}{c}{\textbf{Based on}} & \multicolumn{1}{c}{\textbf{Full train info}} \\
\midrule NAS-Bench-101 & 423k & \cmark & & \xmark \\
NAS-Bench-201 & 6k & \cmark & & \cmark \\
NAS-Bench-301 & $10^{18}$ & \cmark & DARTS & \xmark \\
NAS-Bench-NLP & $10^{53}$ & \xmark & & \xmark \\
NAS-Bench-ASR & 8k & \cmark & & \cmark \\
\midrule
NAS-Bench-111 & 423k & \cmark & NAS-Bench-101 & \cmark \\
NAS-Bench-311 & $10^{18}$ & \cmark & DARTS & \cmark \\
NAS-Bench-NLP11 & $10^{53}$ & \cmark & NAS-Bench-NLP & \cmark \\
\bottomrule
\end{tabular}
\label{tab:benchmarks}
\end{table}

\paragraph{NAS benchmarks.}
NAS-Bench-101~\citep{nasbench}, a tabular NAS benchmark, was created by defining a 
search space of size $423\,624$ unique architectures and then training all architectures from the search space on CIFAR-10 until 108 epochs. However, the train, validation, and test accuracies
are only reported for epochs 4, 12, 36, and 108, and the training, validation, and test 
losses are not reported. NAS-Bench-1shot1~\citep{zela2020bench} defines a subset of 
the NAS-Bench-101 search space that allows one-shot algorithms to be run.
NAS-Bench-201~\citep{nasbench201} contains $15\,625$
architectures, of which $6\,466$ are unique up to isomorphisms.
It comes with full learning curve information on three datasets:
CIFAR-10~\citep{CIFAR10}, CIFAR-100~\citep{CIFAR10},
and ImageNet16-120~\citep{tinyimagenet17}.
Recently, NAS-Bench-201 was extended to NATS-Bench~\citep{natsbench} which
searches over architecture size as well as architecture topology.

Virtually every published NAS method for image classification in the last three years evaluates on the DARTS search space with CIFAR-10~\citep{nb301response}.
The DARTS search space ~\citep{darts} consists of $10^{18}$ neural architectures,
making it computationally prohibitive to create a tabular benchmark. 
To overcome this fundamental limitation and query architectures in this much larger search space, 
NAS-Bench-301~\citep{nasbench301} evaluates various regression models 
trained on a sample of $60\,000$ architectures that is carefully created to cover the whole search 
space. The surrogate models allow users to query the validation accuracy (at epoch 100)
and training time for any of the $10^{18}$ architectures in the DARTS search space.
However, since the surrogates do not predict the entire learning curve, it is not possible to run
multi-fidelity algorithms.

NAS-Bench-NLP~\citep{nasbenchnlp} is a search space for language modeling tasks.
The search space consists of $10^{53}$ architectures, of which
$14\,322$ are evaluated on Penn Tree Bank~\citep{penntreebank}, containing the training, validation,
and test losses/accuracies from epochs 1 to 50. Since only $14\,322$ of $10^{53}$
architectures can be queried, this dataset cannot be directly used for NAS experiments.
NAS-Bench-ASR~\citep{nasbenchasr} is a recent tabular NAS benchmark for speech recognition.
The search space consists of $8\,242$ architectures with full learning curve information.
For an overview of NAS benchmarks, see Table~\ref{tab:benchmarks}.

\section{Creating Surrogate Benchmarks with Learning Curves} \label{sec:surrogate}

In this section, we describe our technique to create a surrogate model 
that outputs realistic learning curves, 
and then we apply this technique to create NAS-Bench-111, NAS-Bench-311, and 
NAS-Bench-NLP11. Our technique applies to any type of learning curve,
including train/test losses and accuracies.
For simplicity, the following presentation assumes validation accuracy learning curves.

\subsection{General Technique} \label{sec:technique} 

\begin{figure}
	\centering
	\includegraphics[width=0.8\textwidth]{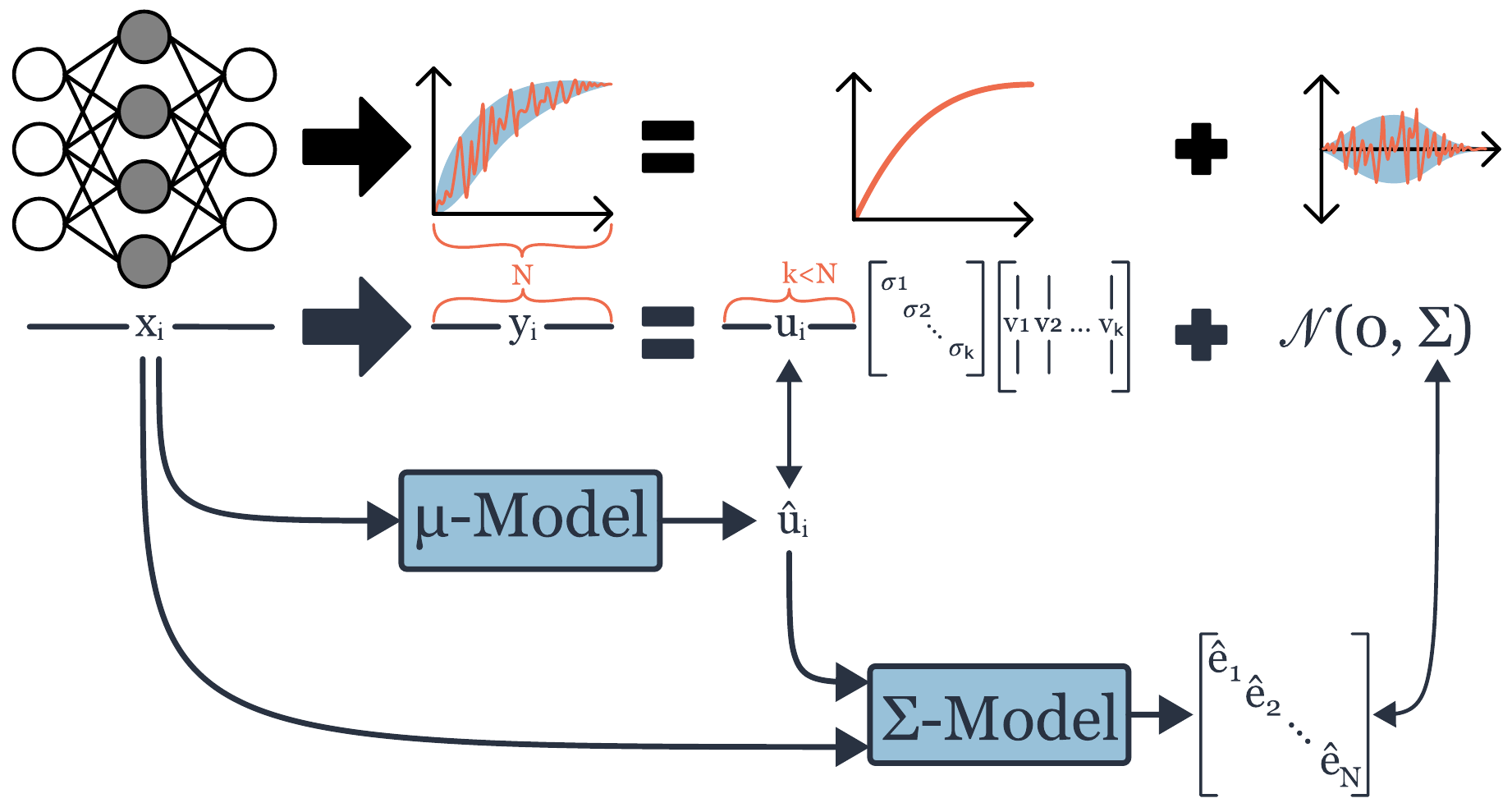}
	\caption{A summary of our approach to create surrogate benchmarks that output realistic
	learning curves. Compression and decompression functions are learned using the training set
	of learning curves (in the figure, SVD is shown, but a VAE can also be used).
	The compression also helps to de-noise the learning curves.
	A model ($\mu$-model) is trained to predict the compressed (de-noised) learning curves
	given the architecture encoding. A separate model ($\Sigma$-model) is trained to predict each
	learning curve's noise distribution, given the architecture encoding and predicted compressed 
	learning curve. A realistic learning curve can then be outputted by decompressing the predicted
	learning curve and sampling noise from the noise distribution.}
	\label{fig:summary}
\end{figure}

Given a search space $\D$,
let $(\vec{x}_i, \vec{y}_i) \sim \D$ denote one datapoint, where 
$\vec{x}_i \in \R^d$ 
is the architecture encoding (e.g., one-hot adjacency matrix~\citep{nasbench, white2020study}), 
and $\vec{y}_i \in [0, 1]^{\Emax}$ 
is a learning curve of validation accuracies drawn from a distribution $Y(\vec x_i)$
based on training the architecture for $\Emax$ epochs on a fixed training 
pipeline with a random initial seed.
Without loss of generality, each learning curve $\vec{y}_i$ can be decomposed into 
two parts: one part that is deterministic and depends only on the architecture encoding, 
and another part that is based on the inherent noise in the architecture 
training pipeline.
Formally, $\vec{y}_i = \E[Y(\vec x_i)] + \vec{\epsilon}_i$,
where $\E[Y(\vec x_i)]\in [0, 1]^{\Emax}$ is fixed and depends only on $\vec{x}_i$,
and $\vec{\epsilon}_i\in [0, 1]^{\Emax}$ comes from a noise distribution
$Z_i$ with expectation 0 for all epochs.
In practice, $\E[Y(\vec x_i)]$ can be estimated by averaging a large set of learning
curves produced by training architecture $\vec{x}_i$ with different initial seeds. 
We represent such an estimate as $\bar{\vec y}_i$. 

Our goal is to create a surrogate model that takes as input any architecture encoding $\vec{x}_i$ 
and outputs a distribution of learning curves that mimics the ground
truth distribution. We assume that we are given two datasets, $\Dtr$ and $\Dte$, of
architecture and learning curve pairs. We use $\Dtr$ (often size $>10\,000$) 
to train the surrogate, and we use $\Dte$ for evaluation. 
We describe the process of creating $\Dtr$ and $\Dte$ for specific search spaces in the next section.
In order to predict a learning curve distribution for each architecture,
we split up our approach into two separate processes: 
we train a model $f: \R^d \to [0,1]^{\Emax}$ to predict the deterministic part of the learning curve, $\bar{\vec y}_i$, 
and we train a noise model $p_\phi (\vec \epsilon \mid \bar{\vec y}, \vec x)$,
parameterized by $\phi$, to simulate the random draws from $Z_i$.

\paragraph{Surrogate model training.}
Training a model $f$ to predict mean learning curves is a challenging task,
since the training datapoints $\Dtr$ consist only of a single (or few) noisy learning curve(s) $\vec{y}_i$ for 
each $\vec{x}_i$.
Furthermore, $\Emax$ is typically length 100 or larger, 
meaning that $f$ must predict a high-dimensional output.
We propose a technique to help with both of these
challenges: we use the training data to learn compression and decompression functions 
$c_k:[0, 1]^{\Emax}\rightarrow [0, 1]^k$ and $d_k :[0, 1]^k\rightarrow [0, 1]^{\Emax}$, 
respectively, for $k \ll \Emax$. 
The surrogate is trained to predict \emph{compressed} learning curves
$c_k\left(\vec{y}_i\right)$ of size $k$ from the corresponding architecture encoding $\vec x_i$, and then each prediction can be reconstructed to a full
learning curve using $d_k$. 
A good compression model should not only cause the surrogate prediction to become
significantly faster and simpler, but should also reduce
the noise in the learning curves, since it would only save the most important information in the compressed representations. 
That is, $(d_k \circ c_k) \left(\vec{y}_i\right)$ 
should be a less noisy version of $\vec{y}_i$. Therefore, models trained on $c_k \left(\vec y_i\right)$ tend to have better generalization ability and do not overfit to the noise
in individual learning curves.

\begin{figure}[t]
	\centering
	\includegraphics[width=0.32\textwidth]{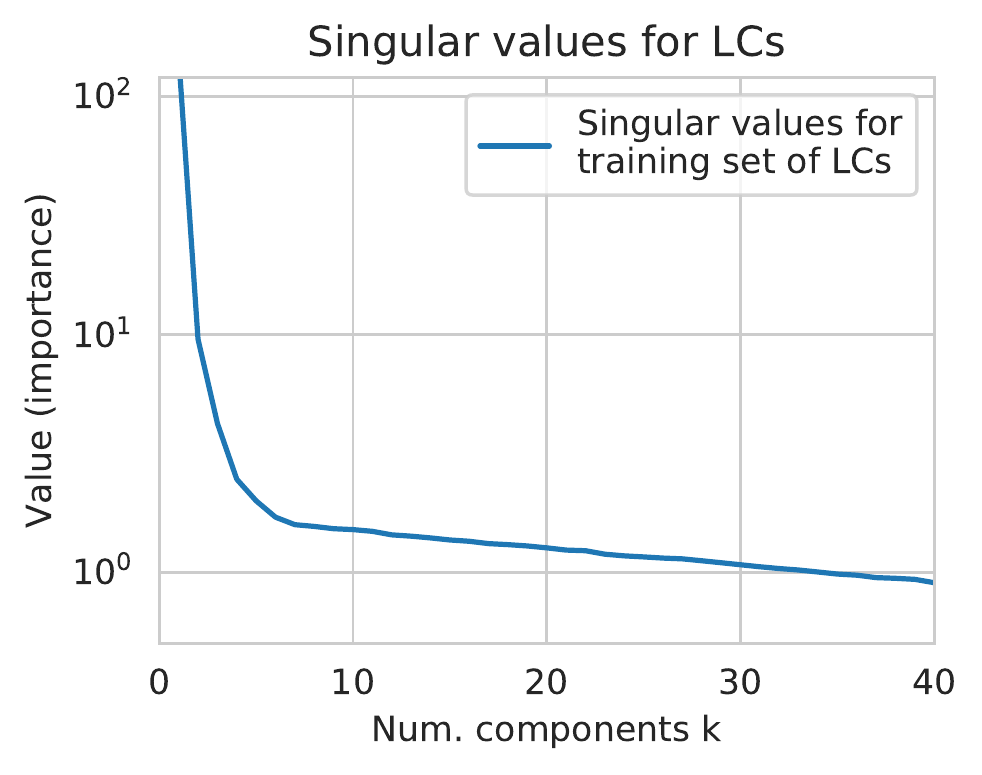}
	\includegraphics[width=0.32\textwidth]{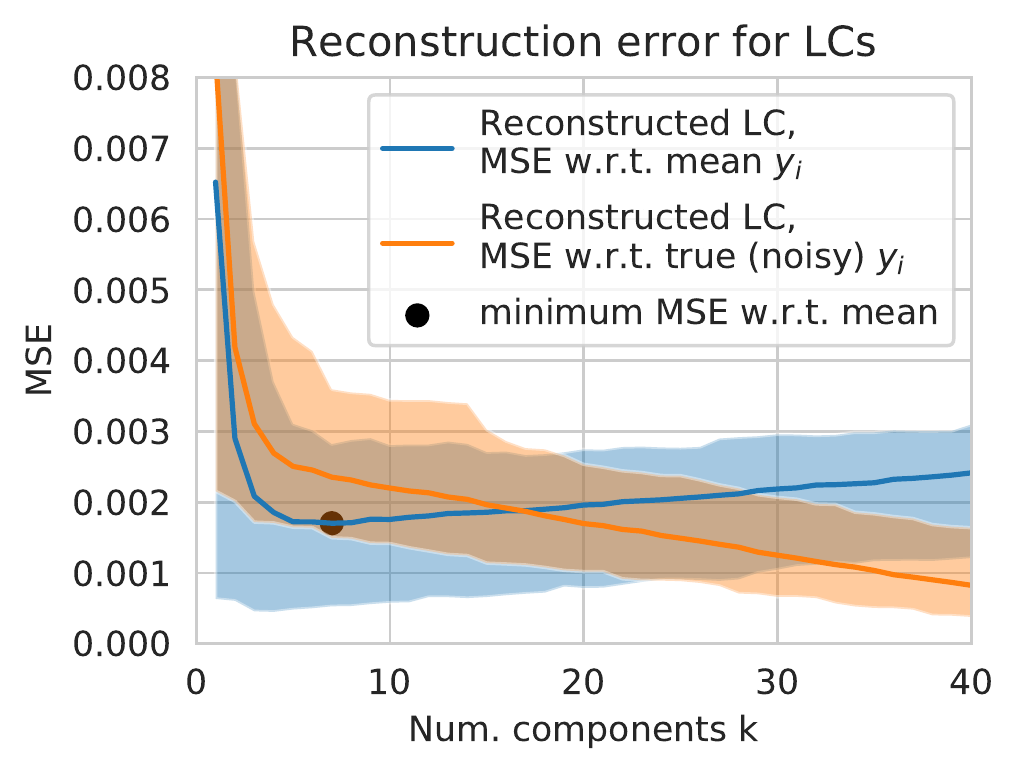}
	\includegraphics[width=0.32\textwidth]{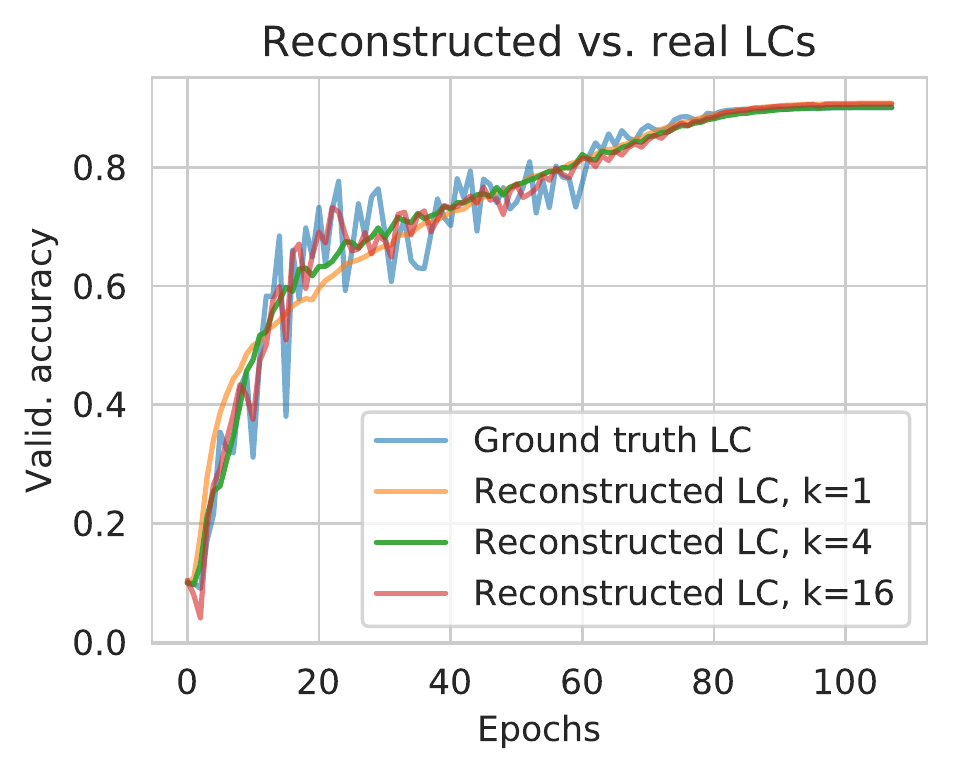}
	\caption{
	The singular values from the SVD decomposition of the learning curves (LC) (left).
	The MSE of a reconstructed LC, showing that $k=6$ is closest to the
	true mean LC, while larger values of $k$ overfit to the noise of the LC (middle).
	An LC reconstructed using different values of $k$ (right).
	}
	\label{fig:svd}
\end{figure}

We test two compression techniques: singular 
value decomposition (SVD)~\citep{golub1965calculating} and variational autoencoders 
(VAEs)~\citep{kingma2013auto}, and we show later that SVD performs better.
We give the details of SVD here and describe the VAE
compression algorithm in Appendix~\ref{app:surrogate}.
Formally, we take the singular value decomposition of a matrix 
$S$
of dimension 
$(|\Dtr|, \Emax)$ created by stacking together the learning curves from all architectures in $\Dtr$.
Performing the truncated SVD on the learning curve matrix $S$ allow us to create functions $c_k$ and $d_k$ that correspond to the optimal linear compression of $S$.
In Figure~\ref{fig:svd} (left), we see that for architectures in the NAS-Bench-101 search
space, there is a steep dropoff of importance after the first six singular values, 
which intuitively means that most of the information for each learning curve is contained in 
its first six singular values. In Figure~\ref{fig:svd} (middle), we compute the mean squared 
error (MSE) of the reconstructed learning curves $(d_k \circ c_k) \left(\vec{y}_i\right)$
compared to a test set of ground truth learning curves averaged over several initial seeds
(approximating $\E[ Y(\vec x_i)]$). The lowest MSE is achieved at $k=6$,
which implies that $k=6$ is the value where the compression function minimizes 
reconstruction error without overfitting to the noise of individual learning curves. 
We further validate this in Figure~\ref{fig:svd} (right)
by plotting $(d_k \circ c_k) \left(\vec{y}_i\right)$ for different values of $k$.
Now that we have compression and decompression functions,
we train a surrogate model with $\vec{x}_i$
as features and $c_k\left(\vec{y}_i\right)$ as the label, for architectures in $\Dtr$.
We test LGBoost~\citep{ke2017lightgbm}, XGBoost~\citep{chen2016xgboost},
and MLPs for the surrogate model.

\paragraph{Noise modeling.}
The final step for creating a realistic surrogate benchmark is to add a noise model,
so that the outputs are \emph{noisy} learning curves. 
We first create a new dataset of predicted $\vec{\epsilon}_i$ values, which we call residuals,
by subtracting the reconstructed mean learning curves from the real learning curves in
$\Dtr$. That is, $\hat {\vec \epsilon}_i = \vec y_i - (d_k \circ c_k) \left(\vec{y}_i\right)$ is the residual for the $i$th learning curve.
Since the training data only contains one (or few) learning curve(s) per 
architecture $\vec x_i$, it is not possible to accurately estimate the distribution 
$Z_i$ for each architecture without making further assumptions.
We assume that the noise comes from an isotropic Gaussian distribution, and we consider two other noise assumptions: \emph{(1)} the noise distribution is the same for all
architectures (to create baselines, or for very homogeneous search spaces), 
and \emph{(2)} for each architecture, the noise in a small window of epochs are iid.
In Appendix~\ref{app:surrogate}, we estimate the extent to which all of these assumptions hold true.
Assumption \emph{(1)} suggests two potential (baseline) noise models: 
\emph{(i)} a simple sample standard deviation statistic, 
$\vec \sigma \in \R_+^{\Emax}$, where 
\begin{equation*}
 \sigma_j = \sqrt{\frac{1}{|\Dtr|-1} \sum_{i=1}^{|\Dtr|} \hat \epsilon_{i,j}^2}.   
\end{equation*}
To sample the noise using this model, we 
sample from $\mathcal{N}(\vec 0, \text{diag}(\vec \sigma))$. \emph{(ii)} The second model is a Gaussian kernel density estimation (GKDE) 
model~\citep{scott2015multivariate} trained on the residuals to create a 
multivariate Gaussian kernel density estimate for the noise.
Assumption \emph{(2)} suggests one potential noise model: 
\emph{(iii)} a model that is trained to estimate the distribution of the 
noise over a window of epochs of a specific architecture. For each architecture and each
epoch, the model is trained to estimate the sample standard deviation of the residuals 
within a window centered around that epoch.

See Figure~\ref{fig:summary} for a summary of the entire surrogate creation method
(assuming SVD).
Throughout this section, we described two compression methods (SVD and VAE), three surrogate models (LGB, XGB, and MLP), and three noise models (stddev, GKDE, and sliding window), for a total of eighteen disctinct approaches. In the next section, we describe the methods we will use to evaluate these approaches.

\subsection{Surrogate Benchmark Evaluation} \label{subsec:evaluation}

The performance of a surrogate benchmark depends on factors such as the size of the training set of architecture + learning curve pairs, the length of the learning curves, and the extent to which the distribution of learning curves satisfy the noise assumptions discussed in the previous section. 
It is important to thoroughly evaluate surrogate benchmarks before their use in NAS research, and so we present a number of different methods for evaluation, which we will use to evaluate the surrogate benchmarks we create in the next section.

We evaluate both the predicted mean learning curves and 
the predicted noise distributions using held-out test sets $\Dte$.
To evaluate the mean learning curves, we compute the coefficient of 
determination ($R^2$)~\citep{wright1921correlation} 
and Kendall Tau (KT) rank correlation~\citep{kendall1938new} between the set of predicted mean learning curves and the set of true learning curves, both at the final epoch and averaged 
over all epochs. 
To measure KT rank correlation for a specific epoch $n$, we find the number of concordant, $P$, and discordant, $Q$, pairs of predicted and true learning curve values for that epoch. The number of concordant pairs is given by the number of pairs, $((\hat{y}_{i,n}, y_{i,n}), (\hat{y}_{j,n}, y_{j,n}))$, where either both $\hat{y}_{i,n} > \hat{y}_{j,n}$ and $y_{i,n} > y_{j,n}$, or both $\hat{y}_{i,n} < \hat{y}_{j,n}$ and $y_{j,n} < y_{i,n}$. We can then calculate $KT = \frac{P - Q} {P + Q}$. 
While these metrics can be used to compare surrogate predictions and have been
used in prior work~\citep{nasbench301}, the rank correlation values are affected by the inherent
noise in the learning curves - even an oracle would have $R^2$ and KT values smaller than 1.0 because architecture training is noisy.
In order to give reference values, we also compute the KT of the set of true learning curves averaged over 2, 3, or 4 independent random seeds. 

The next metric to be used in Section \ref{subsec:creation} is the Kullback Leibler (KL) divergence between the
ground truth distribution of noisy learning curves, and the predicted distribution of 
noisy learning curves on a test set. Since we can only estimate the ground truth distribution, 
we assume the ground truth is an isotropic Gaussian distribution.
Then we measure the KL divergence between the true and predicted learning curves for architecture $i$ by the following formula:

\begin{align*}
    D_{KL}(\vec y_i||\hat{\vec y}_i) = \frac{1}{2 \Emax}\left[\log\frac{|\Sigma_{\hat{\vec y}_i}|}{|\Sigma_{\vec y_i}|} - \Emax + (\boldsymbol{\mu_{\vec y_i}}-\boldsymbol{\mu_{\hat{\vec y}_i}})^T\Sigma_{\hat{\vec y}_i}^{-1}(\boldsymbol{\mu_{\vec y_i}}-\boldsymbol{\mu_{\hat{\vec y}_i}}) + tr\left\{\Sigma_{\hat{\vec y}_i}^{-1}\Sigma_{\vec y_i}\right\}\right]
\end{align*}

where $\Sigma_{\{\vec y_i,\hat{\vec y}_i\}}$ is a diagonal matrix with the entries ${\Sigma_{\{\vec y_i,\hat{\vec y}_i\}}}_{k,k}$ 
representing the sample variance of the $k$-th epoch, and 
$\boldsymbol \mu_{\{\vec y_i,\hat{\vec y}_i\}}$ is the sample mean for either learning curves $\{\vec y_i,\hat{\vec y}_i\}$. 

The final metric is the probability of certain anomalies which we call \emph{spike anomalies}. Even if the KL divergences between the surrogate benchmark distributions and the ground-truth distributions are low, anomalies in the learning curves can still throw off NAS algorithms. For example, there may be anomalies that cause some learning curves to have a much higher maximum validation accuracy than their final validation accuracy.
In order to test for these anomalies, first we compute the largest value $x$ such that there are fewer than 5\% of learning curves whose maximum validation accuracy is $x$ higher than their final validation accuracy, on the real set of learning curves.
Then, we compute the percentage of surrogate learning curves whose maximum validation accuracy is $x$ higher than their final validation accuracy. The goal is for this value to be close to 5\%.

\subsection{Creating NAS-Bench-111, -211, -311, and -NLP11} \label{subsec:creation}

Now we describe the creation of NAS-Bench-111, NAS-Bench-311, and NAS-Bench-NLP11.
We also create NAS-Bench-211 purely to evaluate our technique (since NAS-Bench-201 already has complete learning curves).
Our code and pretrained models are available 
at \url{https://github.com/automl/nas-bench-x11}.
As described above, we test two different compression methods (SVD, VAE), three different
surrogate models (LGB, XGB, MLP), and three different noise models (stddev, GKDE, sliding window)
for a total of eighteen distinct approaches.
See Appendix~\ref{app:surrogate} for a full ablation study 
and Table~\ref{tab:surrogate} for a summary using the best techniques for each
search space. See Figure~\ref{fig:lcs} for a visualization of predicted learning curves from
the test set of each search space using these models.

First, we describe the creation of NAS-Bench-111.
Since the NAS-Bench-101 tabular benchmark~\citep{nasbench} consists only of 
accuracies at epochs 4, 12, 36, and 108 (and without losses), we train a new set
of architectures and save the full learning curves.
Similar to prior work~\citep{eggensperger2015efficient,nasbench301}, we sample a set 
of architectures with good overall coverage while also focusing on the high-performing 
regions exploited by NAS algorithms. Specifically, we sample 861 architectures generated 
uniformly at random, 149 architectures generated by 30 trials of 
BANANAS, local search, and regularized evolution, and all 91 architectures which contain
fewer than five nodes, for a total of 1101 architectures.
We kept our training pipeline as close as possible to the original pipeline.
See Appendix~\ref{app:surrogate} for the full training details. 
We find that SVD-LGB-window and SVD-LGB-GKDE achieve the best performance.
Since the tabular benchmark already exists, 
we can substantially improve the accuracy of our surrogate by using the accuracies
from the tabular benchmark (at epochs 4, 12, 36, 108) as additional features along with the
architecture encoding. This substantially improves the performance of the surrogate, as shown 
in Table~\ref{tab:surrogate}. The large difference between the average KT and last
epoch KT values show that the learning curves are very noisy (which is also evidenced in
Figure~\ref{fig:lcs}).
On a separate test set that contains two seeds evaluated per architecture, we show that the KT values for NAS-Bench-111 are roughly equivalent to those achieved by a 1-seed tabular benchmark (Appendix Table \ref{tab:nb111_kts}).

Next, we create NAS-Bench-311 by using the training data from NAS-Bench-301, 
which consists of $40\,000$ random architectures along with $26\,000$ additional 
architectures generated by evolution~\citep{real2019regularized},
Bayesian optimization~\citep{tpe, bananas, oh2019combinatorial}, 
and one-shot~\citep{darts, pcdarts, gdas, chen2020drnas} techniques in order to achieve
good coverage over the search space.
SVD-LGB-GKDE achieves the best performance. The GKDE model performing well is consistent with prior work that notes DARTS is very homogeneous \citep{yang2019evaluation}.
NAS-Bench-311 achieves average and last epoch KT
values of 0.728 and 0.788, respectively. This is comparable to the last epoch KT of 0.817 reported by 
NAS-Bench-301~\citep{nasbench301}, despite optimizing for the full learning curve rather than only 
the final epoch. Furthermore, our KL divergences in Table~\ref{tab:surrogate} 
surpasses the top value of 16.4 reported by NAS-Bench-301 (for KL divergence, lower is better).
In Appendix Table \ref{tab:nb311_kts}, we show that the mean model in NAS-Bench-311 achieves higher rank correlation even than a set of learning curves averaged over four random seeds, by using a separate test set from the NAS-Bench-301 dataset which evaluates 500 architectures with 5 seeds each.
We also show that the percentage of spike anomalies for real vs.\ surrogate data (defined in the previous section) is 5\% and 7.02\%, respectively.

Finally, we create NAS-Bench-NLP11 by using the NAS-Bench-NLP dataset consisting of
$14\,322$ architectures drawn uniformly at random. 
Due to the extreme size of the search space ($10^{53}$), 
we achieve an average and final epoch KT of 0.449 and 0.416, respectively.
Since there are no architectures trained multiple times on the NAS-Bench-NLP 
dataset~\citep{nasbenchnlp}, we cannot compute the KL divergence or additional metrics.
To create a stronger surrogate, we add the first three epochs
of the learning curve as features in the surrogate.
This improves the average and final epoch KT values to 0.862 and 0.820, respectively.
To use this surrogate, any architecture to be predicted with the surrogate must be trained for 
three epochs. Note that the small difference between the average and last epoch KT values indicates
that the learning curves have very little noise, which can also be seen in Figure~\ref{fig:lcs}.
This technique can be used to improve the performance of other surrogates too, such as NAS-Bench-311. 
For NAS-Bench-311, we find that adding the validation accuracies from the first three epochs as additional epochs, improves the average and final epoch KT from 0.728 and 0.788 
to 0.749 and 0.795, respectively.

We also create NAS-Bench-211 to further evaluate our surrogate creation technique (only for evaluation purposes, since NAS-Bench-201 already has complete learning curves). We train the surrogate on 90\% of architectures from NAS-Bench-201 (14\,062 architectures) and test on the remaining 10\%. SVD-LGB-window achieves the best performance.
The average and final KT values are 0.701 and 0.842, respectively, which is on par with a 1-seed tabular benchmark (see Table \ref{tab:nb211_kts}).
We also show that the percentage of spike anomalies for real vs.\ surrogate data (defined in the previous section) is 5\% and 10.14\%, respectively.

\begin{table} 
\caption{Evaluation of the surrogate benchmarks on test sets.
For NAS-Bench-111 and NAS-Bench-NLP11, we use architecture accuracies
as additional features to improve performance.
} \label{tab:surrogate}
\centering
{\small
\begin{tabular}{@{}l|c|c|c|c|c|c@{}}
\toprule
\multicolumn{1}{l}{\textbf{Benchmark}} & \multicolumn{1}{c}{\textbf{Avg.\ $R^2$}} & \multicolumn{1}{c}{\textbf{Final $R^2$}} & \multicolumn{1}{c}{\textbf{Avg.\ KT}} & \multicolumn{1}{c}{\textbf{Final KT}} & \multicolumn{1}{c}{\textbf{Avg.\ KL}} & \multicolumn{1}{c}{\textbf{Final KL}}\\
\midrule 
NAS-Bench-111 & 0.529 & 0.630 & 0.531 & 0.645 & 2.016 & 1.061 \\
NAS-Bench-111 (w.\ accs) & 0.630 & 0.853 & 0.611 & 0.794 & 1.710 & 0.926 \\
NAS-Bench-311 & 0.779 & 0.800 & 0.728 & 0.788 & 0.905 & 0.600 \\
NAS-Bench-NLP11 & 0.327 & 0.292 & 0.449 & 0.416 & - & - \\
NAS-Bench-NLP11 (w.\ accs) & 0.906 & 0.882 & 0.862 & 0.820 & - & - \\
\bottomrule
\end{tabular}
}
\end{table}

\section{The Power of Learning Curve Extrapolation}\label{sec:lce}


Now we describe a simple framework for converting single-fidelity NAS algorithms to
multi-fidelity NAS algorithms using learning curve extrapolation techniques.
We show that this framework is able to substantially improve the performance of popular 
algorithms such as regularized evolution~\citep{real2019regularized},
BANANAS~\citep{bananas}, and local search~\citep{white2020local, ottelander2020local}.

A single-fidelity algorithm is an algorithm which iteratively chooses an architecture
based on its history, which is then fully trained to $\Emax$ epochs. 
To exploit parallel resources, many single-fidelity algorithms iteratively output several
architectures at a time, instead of just one. In Algorithm~\ref{alg:sf},
we present pseudocode for a generic single-fidelity algorithm. For example,
for local search, \texttt{gen\_candidates} would output the neighbors of the architecture 
with the best accuracy in \texttt{history}.

Our framework makes use of learning curve extrapolation (LCE) 
techniques~\citep{domhan2015speeding,baker2017accelerating} 
to predict the final validation accuracies of all architecture choices
after only training for a small number of epochs. 
See Algorithm~\ref{alg:lce}.
After each iteration of \texttt{gen\_candidates}, only the architectures predicted by \texttt{LCE()} to be
in the top percentage of validation accuracies of \texttt{history} are fully trained.
For example, when the framework is applied
to local search, in each iteration, all neighbors are trained to $E_{\text{few}}$
epochs, and only the most promising architectures are trained up to $\Emax$ epochs.
This simple modification can substantially improve the runtime efficiency of popular NAS algorithms by weeding out unpromising architectures before they are fully trained.  

Any LCE technique can be used, and in our experiments in 
Section~\ref{sec:experiments}, we use
weighted probabilistic modeling (WPM)~\citep{domhan2015speeding} and 
learning curve support vector regressor (LcSVR)~\citep{baker2017accelerating}.
The first technique, WPM~\citep{domhan2015speeding}, is a function that takes a partial
learning curve as input, and then extrapolates it by fitting the learning curve to a set of 
parametric functions, using MCMC to sample the most promising fit.
The second technique, LcSVR~\citep{baker2017accelerating}, is a model-based learning curve extrapolation technique: after generating an initial set of training architectures, a support vector regressor is trained to predict the final validation accuracy from the architecture encoding and partial learning curve.

\noindent\begin{minipage}[t]{\textwidth}
\begin{minipage}[t]{0.45\textwidth}
\begin{algorithm}[H]
    \centering
    \caption{Single-Fidelity Algorithm}\label{alg:sf}
    \footnotesize
    \begin{algorithmic}[1]
        \STATE \texttt{initialize history}
        \STATE \texttt{while} $t<t_{\max}:$
        \STATE \quad \texttt{arches = gen\_candidates(history)}
        \STATE \quad \texttt{accs = train(arches, epoch=$\Emax$)}
        \STATE \quad \texttt{history.update(arches, accs)}
        \STATE \textbf{Return} \texttt{arch} with the highest \texttt{acc}
    \end{algorithmic}
\end{algorithm}
\end{minipage}
\hfill
\begin{minipage}[t]{0.52\textwidth}
\begin{algorithm}[H]
    \centering
    \caption{LCE Framework}\label{alg:lce}
    \footnotesize
    \begin{algorithmic}[1]
        \STATE\texttt{initialize history}
        \STATE \texttt{while} $t<t_{\max}:$
        \STATE\quad\texttt{arches = gen\_candidates(history)}
        \STATE\quad\texttt{accs = train(arches, epoch=$E_{\text{few}}$)}
        \STATE\quad\texttt{sorted\_by\_pred = LCE(arches, accs)}
        \STATE\quad\texttt{arches = sorted\_by\_pred[:top\_n]}
        \STATE\quad\texttt{accs = train(arches, epoch=$\Emax$)}\\
        \STATE\quad\texttt{history.update(arches, accs)}
        \STATE \textbf{Return} \texttt{arch} with the highest \texttt{acc}
    \end{algorithmic}
\end{algorithm}
\end{minipage}
\end{minipage}

\section{Experiments} \label{sec:experiments}

In this section, we benchmark single-fidelity and multi-fidelity NAS algorithms, including
popular existing single-fidelity and multi-fidelity algorithms, as well as algorithms
created using our framework defined in the previous section.
In the experiments, we use our three surrogate benchmarks defined in 
Section~\ref{sec:surrogate}, as well as NAS-Bench-201.


\paragraph{NAS algorithms.} For single-fidelity algorithms, we implemented random search (RS)~\citep{randomnas}, local search (LS)~\citep{white2020local, ottelander2020local},
regularized evolution (REA)~\citep{real2019regularized}, and BANANAS~\citep{bananas}.
For multi-fidelity bandit-based algorithms, we implemented Hyperband (HB)~\citep{hyperband} 
and Bayesian optimization Hyperband (BOHB)~\citep{bohb}. For all methods, we use the original implementation whenever possible. See Appendix~\ref{app:experiments} for a description, implementation details, and hyperparameter details for each method. 
Finally, we use our framework from Section~\ref{sec:lce} to create six new multi-fidelity algorithms: BANANAS, LS, and REA are each augmented using WPM and 
LcSVR. This gives us a total of $12$ algorithms.

\paragraph{Experimental setup.}
For each search space, we run each algorithm for a total wall-clock time that is equivalent
to running 500 iterations of the single-fidelity algorithms for NAS-Bench-111 and NAS-Bench-311, and 100 iterations for NAS-Bench-201 and NAS-Bench-NLP11. For example, the average time to train a NAS-Bench-111 architecture to 108 epochs is roughly $10^3$ seconds, so we set the maximum runtime on NAS-Bench-111 to roughly $5\cdot 10^5$ seconds.
We run 30 trials of each NAS algorithm and compute the mean and standard deviation.


\begin{wrapfigure}{t}{2.0in} 
\begin{minipage}{2.0in}
\centering
\resizebox{2.0in}{!}{\includegraphics{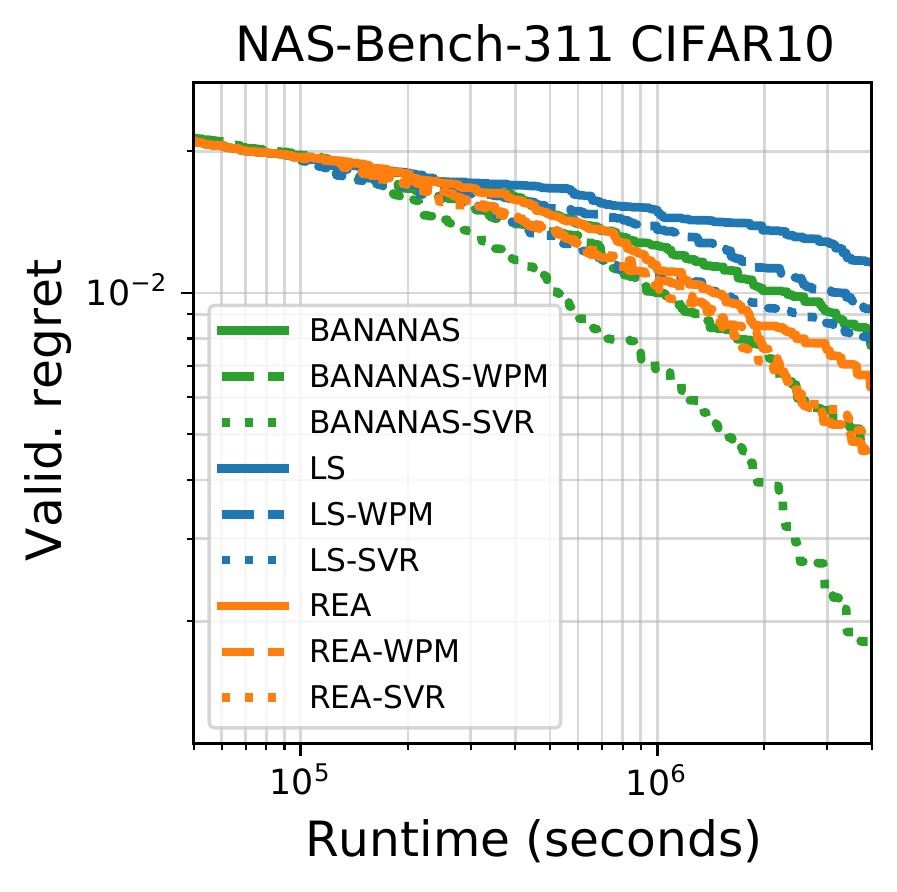}}
\caption{LCE Framework applied to single-fidelity algorithms on NAS-Bench-311. 
}
\label{fig:lce}
\end{minipage}
\end{wrapfigure}

\paragraph{Results.}
We evaluate BANANAS, LS, and REA compared to their augmented WPM and SVR versions
in Figure~\ref{fig:lce} (NAS-Bench-311) and Appendix~\ref{app:experiments} 
(all other search spaces).
Across four search spaces, we see that WPM and SVR improve all algorithms in almost all settings. The improvements are particularly strong for the larger NAS-Bench-111 and
NAS-Bench-311 search spaces.
We also see that for each single-fidelity algorithm, the LcSVR variant often 
outperforms the WPM
variant. This suggests that model-based techniques for extrapolating learning curves are more reliable
than extrapolating each learning curve individually, which has also been noted in prior work~\citep{white2021powerful}.

\begin{figure}
\centering
\includegraphics[width=.32\columnwidth]{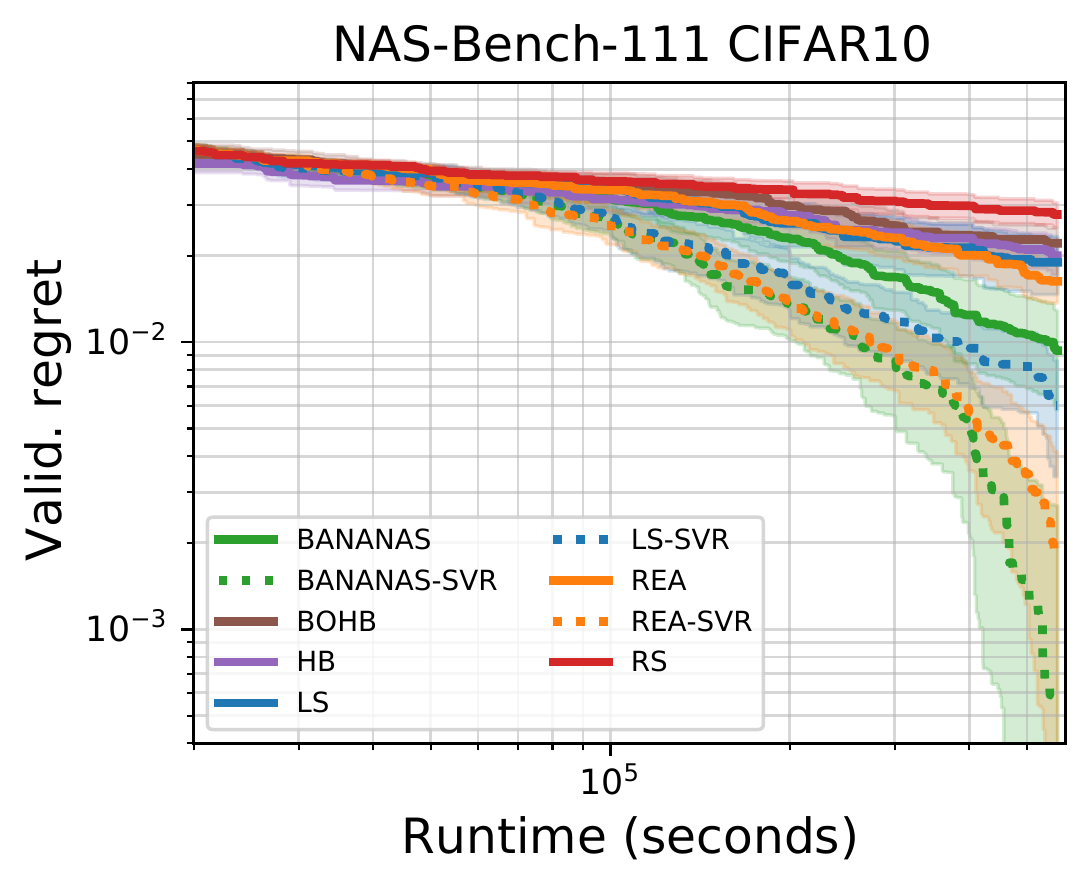}
\includegraphics[width=.32\columnwidth]{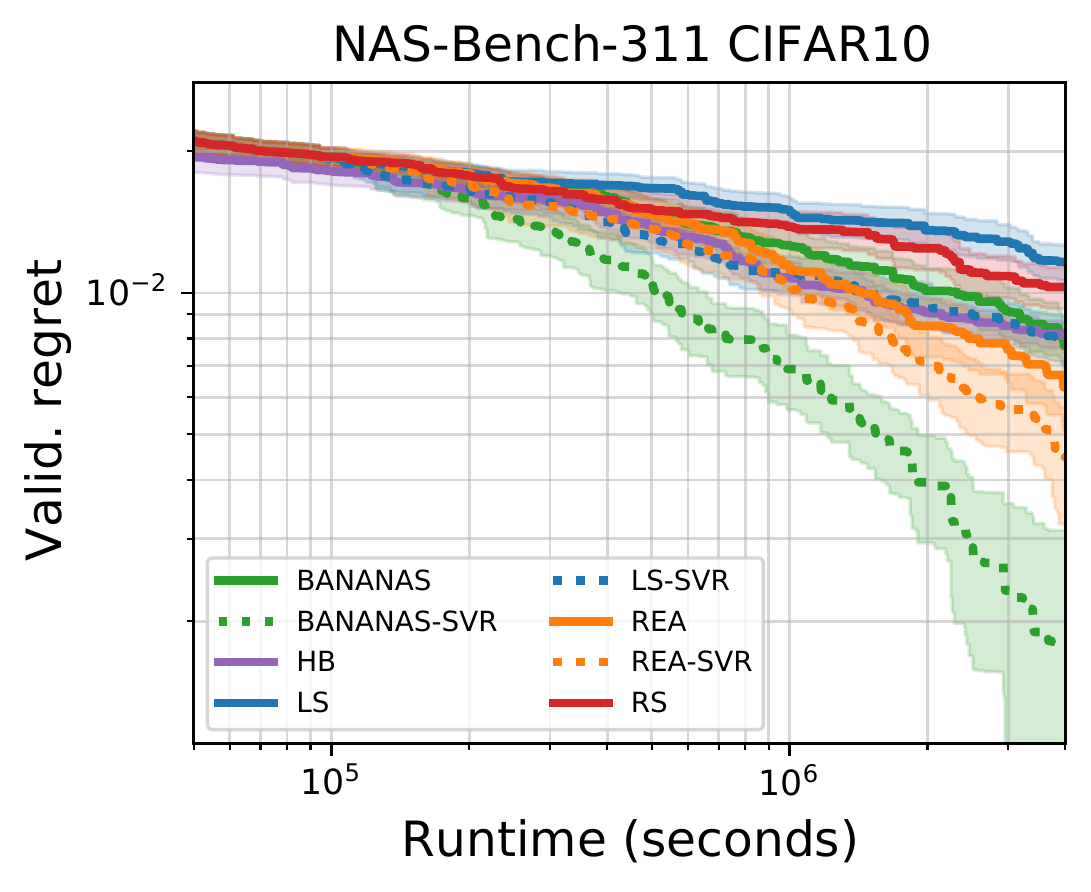}
\includegraphics[width=.32\columnwidth]{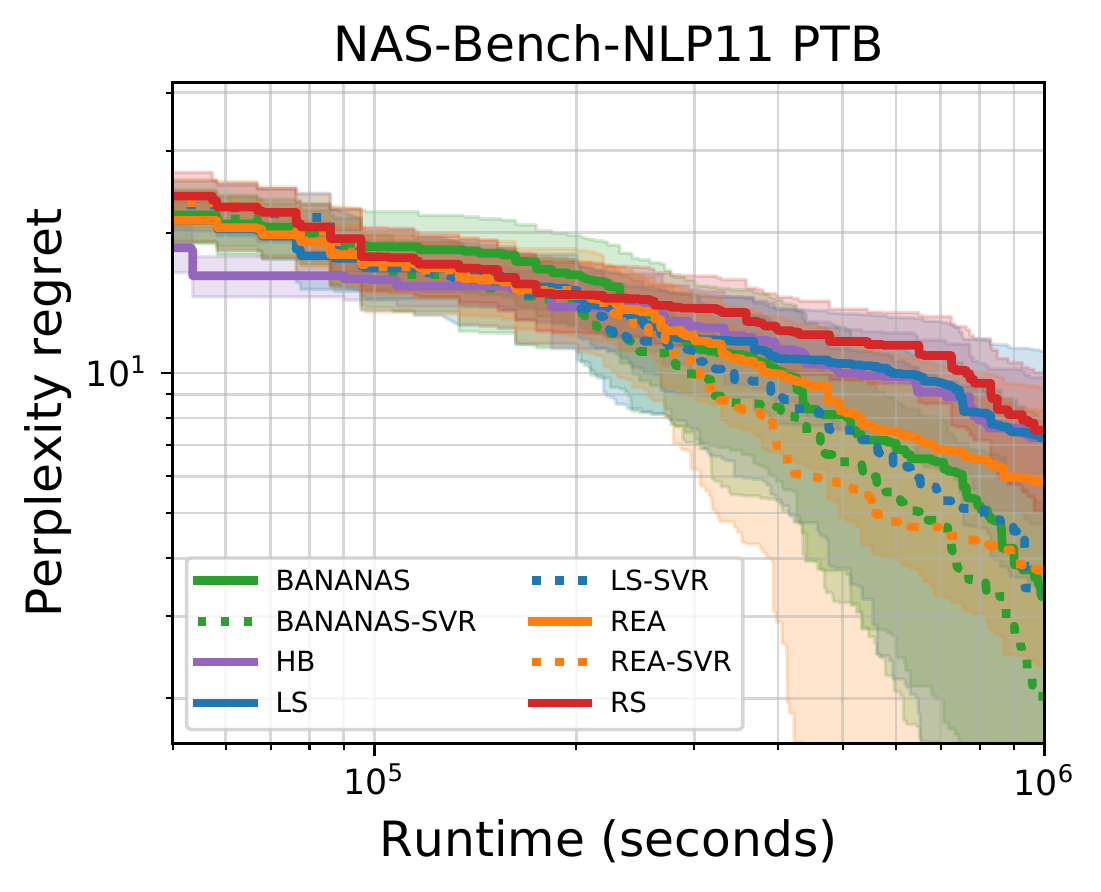}
\includegraphics[width=.32\columnwidth]{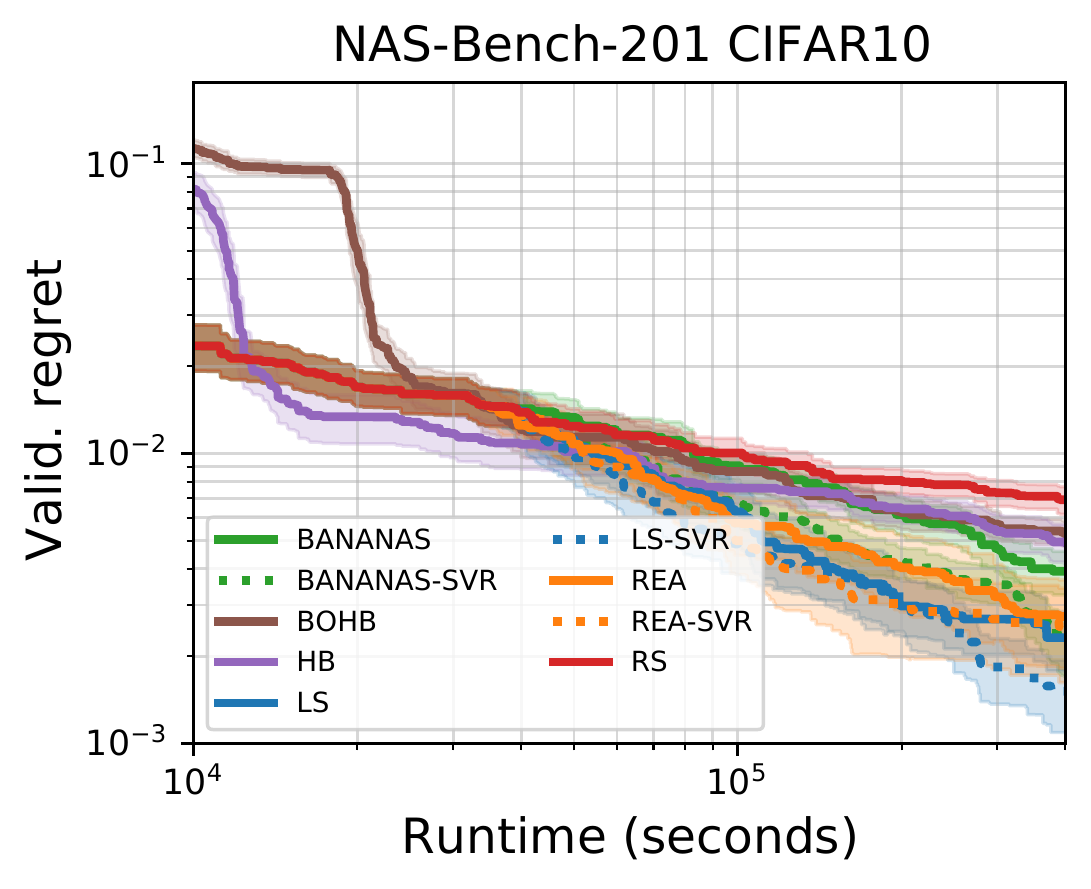}
\includegraphics[width=.32\columnwidth]{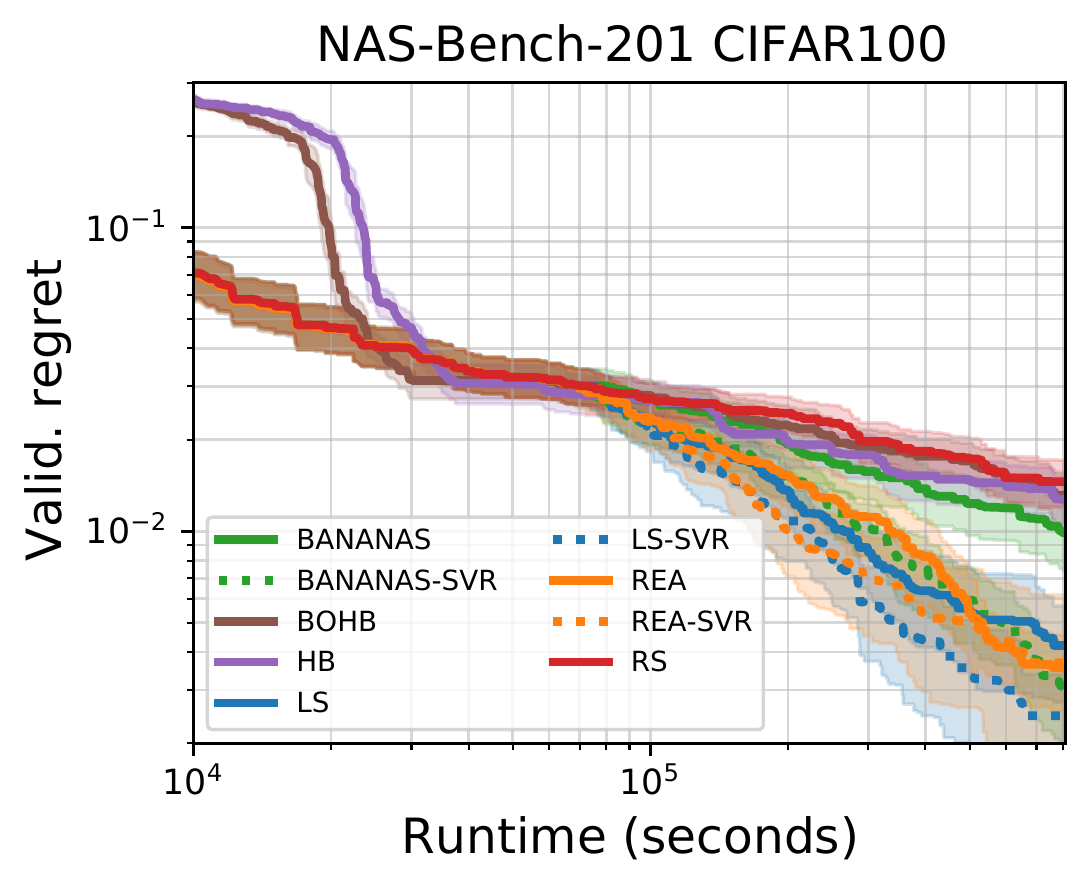}
\includegraphics[width=.32\columnwidth]{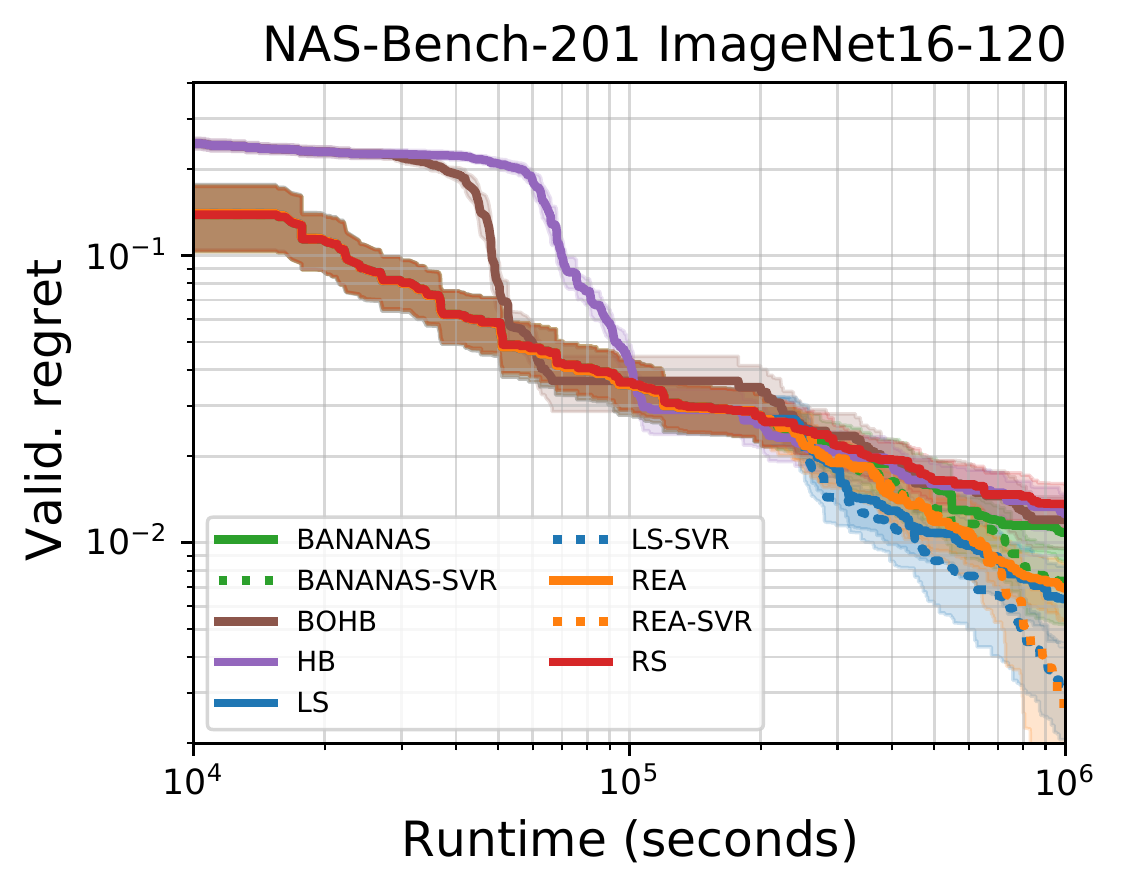}
\caption{NAS results on six different combinations of search spaces and datasets.
For every setting, an SVR augmented method performs best.} 
\label{fig:nas}
\end{figure}

In Figure~\ref{fig:nas}, we compare single- and multi-fidelity algorithms on four search spaces, along with the three SVR-based algorithms from Figure~\ref{fig:lce}. 
Across all search spaces, an SVR-based algorithm is the top-performing algorithm.
Specifically, BANANAS-SVR performs the best on NAS-Bench-111, NAS-Bench-311, and NAS-Bench-NLP11, and LS-SVR performs the best on NAS-Bench-201. 
Note that HB and BOHB may not perform well on search spaces with low correlation 
between the relative rankings of architectures using low fidelities and high fidelities
(such as NAS-Bench-101~\citep{nasbench} and NAS-Bench-201~\citep{nasbench201})
since HB-based methods will predict the final accuracy of partially trained architectures directly from
the last trained accuracy (i.e., extrapolating the learning curve as a constant after 
the last seen accuracy).
On the other hand, SVR-based approaches use a model that can learn more complex relationships between
accuracy at an early epoch vs.\ accuarcy at the final epoch, and are therefore more robust to 
this type of search space.

In Appendix~\ref{app:experiments},
we perform an ablation study on the epoch at which the SVR and WPM methods start extrapolating in our
framework (i.e., we ablate $E_{\text{few}}$ from Algorithm~\ref{alg:lce}).
We find that for most search spaces, running SVR and WPM
based NAS methods by starting the LCE at roughly $20\%$ of the total number of epochs performs the best.
Any earlier, and there is not enough information to accurately extrapolate the learning curve.
Any later, and the LCE sees diminishing returns because less time is saved by early stopping the
training.

\section{Societal Impact} \label{sec:impact}
%
Our hope is that our work will make it quicker and easier for researchers to run fair experiments and 
give reproducible conclusions.
In particular, the surrogate benchmarks allow AutoML
researchers to develop NAS algorithms directly on a CPU, as opposed 
to using a GPU, which may decrease the carbon emissions from GPU-based NAS 
research~\citep{patterson2021carbon, hao2019training}.
In terms of our proposed NAS speedups, these techniques are a level of abstraction away 
from real applications, but they can indirectly impact the broader society.
For example, this work may facilitate the creation of new high-performing NAS techniques,
which can then be used to improve various deep learning methods, both beneficial
(e.g.\ algorithms that reduce $\text{CO}_2$ emissions), or harmful 
(e.g.\ algorithms that produce deep fakes).

\section{Conclusions, Limitations, and Guidelines} \label{sec:conclusion}
In this work, we released three benchmarks for neural architecture search based on three popular 
search spaces, which substantially improve the capability of existing benchmarks due to the availability 
of the full learning curve for train/validation/test loss and accuracy for each architecture.
Our techniques to generate these benchmarks, which includes singular value decomposition of the learning
curve and noise modeling, can be used to model the full learning curve for future surrogate NAS benchmarks as well.

Furthermore, we demonstrated the power of the full learning curve information by introducing a framework
that converts single-fidelity NAS algorithms into multi-fidelity NAS algorithms that make use of
learning curve extrapolation techniques. This framework improves the performance
of recent popular single-fidelity algorithms which claimed to be state-of-the-art upon release.



While we believe our surrogate benchmarks will help advance scientific research in NAS,
a few guidelines and limitations are important to keep in mind.
As with prior surrogate benchmarks~\citep{nasbench301}, we give the following two caveats.
\emph{(1)} We discourage evaluating NAS methods that use the same internal techniques as those used
in the surrogate model. For example, any NAS method that makes use of variational autoencoders
or XGBoost should not be benchmarked using our VAE-XGB surrogate benchmark.
\emph{(2)} As the surrogate benchmarks are likely to evolve as new training data is added, or as
better techniques for training a surrogate are devised, we recommend reporting the surrogate
benchmark version number whenever running experiments.
\emph{(3)} When creating new surrogate benchmarks, before use in NAS, it is important to give a thorough evaluation such as the evaluation methods described in Section \ref{subsec:evaluation}.

We also note the following strengths and limitations for specific benchmarks.
Our NAS-Bench-111 surrogate benchmark gives strong mean performance even with just $1\,101$
architectures used as training data, due to the existence of the four extra validation
accuracies from NAS-Bench-101 that can be used as additional features. We hope that all future tabular
benchmarks will save the full learning curve information from the start so that the creation of
an after-the-fact extended benchmark is not necessary.

Although all of our surrogates had strong mean performance, the noise model was strongest only for NAS-Bench-311, which had a training dataset of size 60k, as evidenced especially because of the rate of spike anomalies, described in Section \ref{subsec:evaluation}. This can be mitigated by adding more training data.

Since the NAS-Bench-NLP11 surrogate benchmark achieves significantly stronger performance when the
accuracy of the first three epochs are added as features, we recommend using this benchmark by
training architectures for three epochs before querying the surrogate. Therefore, benchmarking NAS
algorithms are slower than for NAS-Bench-111 and NAS-Bench-311, but 
NAS-Bench-NLP11 still offers a $15\times$
speedup compared to a NAS experiment without this benchmark. We still release the version
of NAS-Bench-NLP11 that does not use the first three accuracies as features but with a 
warning that the observed NAS trends may differ from the true NAS trends. We expect that the
performance of these benchmarks will improve over time, as data for more 
trained architectures become available.

\begin{ack}
Work done while the first three authors were working at Abacus.AI.
FH acknowledges support by the European Research Council (ERC) under the European Union Horizon 2020 research and innovation programme through grant no. 716721, BMBF grant DeToL, and TAILOR, a project funded by EU Horizon 2020 research and innovation programme under GA No 952215.
We thank Kaicheng Yu for his discussions with this project.
\end{ack}

\newpage
\bibliography{main}

\begin{thebibliography}{10}

\bibitem{baker2017accelerating}
Bowen Baker, Otkrist Gupta, Ramesh Raskar, and Nikhil Naik.
\newblock Accelerating neural architecture search using performance prediction.
\newblock In {\em ICLR Workshop}, 2018.

\bibitem{bender2018understanding}
Gabriel Bender, Pieter-Jan Kindermans, Barret Zoph, Vijay Vasudevan, and Quoc
  Le.
\newblock Understanding and simplifying one-shot architecture search.
\newblock In {\em ICML}, 2018.

\bibitem{tpe}
James~S Bergstra, R{\'e}mi Bardenet, Yoshua Bengio, and Bal{\'a}zs K{\'e}gl.
\newblock Algorithms for hyper-parameter optimization.
\newblock In {\em NeurIPS}, 2011.

\bibitem{chandrashekaran2017speeding}
Akshay Chandrashekaran and Ian~R Lane.
\newblock Speeding up hyper-parameter optimization by extrapolation of learning
  curves using previous builds.
\newblock In {\em ECML-PKDD}, 2017.

\bibitem{chen2016xgboost}
Tianqi Chen and Carlos Guestrin.
\newblock Xgboost: A scalable tree boosting system.
\newblock In {\em Proceedings of the 22nd acm sigkdd international conference
  on knowledge discovery and data mining}, pages 785--794, 2016.

\bibitem{chen2020drnas}
Xiangning Chen, Ruochen Wang, Minhao Cheng, Xiaocheng Tang, and Cho-Jui Hsieh.
\newblock Drnas: Dirichlet neural architecture search.
\newblock In {\em ICLR}, 2021.

\bibitem{tinyimagenet17}
Patryk Chrabaszcz, Ilya Loshchilov, and Frank Hutter.
\newblock A downsampled variant of imagenet as an alternative to the cifar
  datasets.
\newblock In {\em arXiv:1707.08819}, 2017.

\bibitem{domhan2015speeding}
Tobias Domhan, Jost~Tobias Springenberg, and Frank Hutter.
\newblock Speeding up automatic hyperparameter optimization of deep neural
  networks by extrapolation of learning curves.
\newblock In {\em IJCAI}, 2015.

\bibitem{natsbench}
Xuanyi Dong, Lu~Liu, Katarzyna Musial, and Bogdan Gabrys.
\newblock Nats-bench: Benchmarking nas algorithms for architecture topology and
  size.
\newblock In {\em PAMI}, 2021.

\bibitem{gdas}
Xuanyi Dong and Yi~Yang.
\newblock Searching for a robust neural architecture in four gpu hours.
\newblock In {\em CVPR}, 2019.

\bibitem{nasbench201}
Xuanyi Dong and Yi~Yang.
\newblock Nas-bench-201: Extending the scope of reproducible neural
  architecture search.
\newblock In {\em ICLR}, 2020.

\bibitem{eggensperger2015efficient}
Katharina Eggensperger, Frank Hutter, Holger Hoos, and Kevin Leyton-Brown.
\newblock Efficient benchmarking of hyperparameter optimizers via surrogates.
\newblock In {\em AAAI}, 2015.

\bibitem{nas-survey}
Thomas Elsken, Jan~Hendrik Metzen, and Frank Hutter.
\newblock Neural architecture search: A survey.
\newblock In {\em JMLR}, 2019.

\bibitem{bohb}
Stefan Falkner, Aaron Klein, and Frank Hutter.
\newblock Bohb: Robust and efficient hyperparameter optimization at scale.
\newblock In {\em ICML}, 2018.

\bibitem{gargiani2019probabilistic}
Matilde Gargiani, Aaron Klein, Stefan Falkner, and Frank Hutter.
\newblock Probabilistic rollouts for learning curve extrapolation across
  hyperparameter settings.
\newblock {\em arXiv preprint arXiv:1910.04522}, 2019.

\bibitem{golub1965calculating}
Gene Golub and William Kahan.
\newblock Calculating the singular values and pseudo-inverse of a matrix.
\newblock {\em Journal of the Society for Industrial and Applied Mathematics,
  Series B: Numerical Analysis}, 2(2):205--224, 1965.

\bibitem{hao2019training}
Karen Hao.
\newblock Training a single ai model can emit as much carbon as five cars in
  their lifetimes.
\newblock {\em MIT Technology Review}, 2019.

\bibitem{hu2019forwardnas}
Hanzhang Hu, John Langford, Rich Caruana, Saurajit Mukherjee, Eric Horvitz, and
  Debadeepta Dey.
\newblock Efficient forward architecture search.
\newblock In {\em NeurIPS}, 2019.

\bibitem{huang2020asymptotically}
Yimin Huang, Yujun Li, Hanrong Ye, Zhenguo Li, and Zhihua Zhang.
\newblock An asymptotically optimal multi-armed bandit algorithm and
  hyperparameter optimization.
\newblock {\em arXiv preprint arXiv:2007.05670}, 2020.

\bibitem{k2016multifidelity}
Kirthevasan Kandasamy, Gautam Dasarathy, Junier~B. Oliva, Jeff Schneider, and
  Barnabas Poczos.
\newblock Multi-fidelity gaussian process bandit optimisation.
\newblock In {\em NeurIPS}, 2016.

\bibitem{k2017multifidelity}
Kirthevasan Kandasamy, Gautam Dasarathy, Jeff Schneider, and Barnabas Poczos.
\newblock Multi-fidelity bayesian optimisation with continuous approximations.
\newblock In {\em JMLR}, 2017.

\bibitem{nasbot}
Kirthevasan Kandasamy, Willie Neiswanger, Jeff Schneider, Barnabas Poczos, and
  Eric~P Xing.
\newblock Neural architecture search with bayesian optimisation and optimal
  transport.
\newblock In {\em NeurIPS}, 2018.

\bibitem{ke2017lightgbm}
Guolin Ke, Qi~Meng, Thomas Finley, Taifeng Wang, Wei Chen, Weidong Ma, Qiwei
  Ye, and Tie-Yan Liu.
\newblock Lightgbm: A highly efficient gradient boosting decision tree.
\newblock In {\em NeurIPS}, 2017.

\bibitem{kendall1938new}
Maurice~G Kendall.
\newblock A new measure of rank correlation.
\newblock {\em Biometrika}, 30(1/2):81--93, 1938.

\bibitem{kingma2013auto}
Diederik~P Kingma and Max Welling.
\newblock Auto-encoding variational bayes.
\newblock In {\em ICLR}, 2014.

\bibitem{kitano1990designing}
Hiroaki Kitano.
\newblock Designing neural networks using genetic algorithms with graph
  generation system.
\newblock {\em Complex systems}, 4(4):461--476, 1990.

\bibitem{klein17fast}
Aaron Klein, Stefan Falkner, Simon Bartels, Philipp Hennig, and Frank Hutter.
\newblock {Fast Bayesian Optimization of Machine Learning Hyperparameters on
  Large Datasets}.
\newblock In {\em AISTATS}, 2017.

\bibitem{lcnet}
Aaron Klein, Stefan Falkner, Jost~Tobias Springenberg, and Frank Hutter.
\newblock Learning curve prediction with bayesian neural networks.
\newblock In {\em ICLR}, 2017.

\bibitem{abohb}
Aaron Klein, Louis Tiao, Thibaut Lienart, Cedric Archambeau, and Matthias
  Seeger.
\newblock Model-based asynchronous hyperparameter and neural architecture
  search.
\newblock {\em arXiv preprint arXiv:2003.10865}, 2020.

\bibitem{nasbenchnlp}
Nikita Klyuchnikov, Ilya Trofimov, Ekaterina Artemova, Mikhail Salnikov, Maxim
  Fedorov, and Evgeny Burnaev.
\newblock Nas-bench-nlp: neural architecture search benchmark for natural
  language processing.
\newblock {\em arXiv preprint arXiv:2006.07116}, 2020.

\bibitem{CIFAR10}
Alex Krizhevsky.
\newblock Learning multiple layers of features from tiny images.
\newblock Technical report, University of Toronto, 2009.

\bibitem{li2018massively}
Liam Li, Kevin Jamieson, Afshin Rostamizadeh, Ekaterina Gonina, Moritz Hardt,
  Benjamin Recht, and Ameet Talwalkar.
\newblock A system for massively parallel hyperparameter tuning.
\newblock In {\em MLSys Conference}, 2020.

\bibitem{li2020geometry}
Liam Li, Mikhail Khodak, Maria-Florina Balcan, and Ameet Talwalkar.
\newblock Geometry-aware gradient algorithms for neural architecture search.
\newblock In {\em Proceedings of the International Conference on Learning
  Representations (ICLR)}, 2021.

\bibitem{randomnas}
Liam Li and Ameet Talwalkar.
\newblock Random search and reproducibility for neural architecture search.
\newblock In {\em UAI}, 2019.

\bibitem{hyperband}
Lisha Li, Kevin Jamieson, Giulia DeSalvo, Afshin Rostamizadeh, and Ameet
  Talwalkar.
\newblock Hyperband: A novel bandit-based approach to hyperparameter
  optimization.
\newblock In {\em JMLR}, 2018.

\bibitem{lindauer2019best}
Marius Lindauer and Frank Hutter.
\newblock Best practices for scientific research on neural architecture search.
\newblock In {\em JMLR}, 2020.

\bibitem{liu2017progressive}
Chenxi Liu, Barret Zoph, Maxim Neumann, Jonathon Shlens, Wei Hua, Li-Jia Li,
  Li~Fei-Fei, Alan Yuille, Jonathan Huang, and Kevin Murphy.
\newblock Progressive neural architecture search.
\newblock In {\em ECCV}, 2018.

\bibitem{darts}
Hanxiao Liu, Karen Simonyan, and Yiming Yang.
\newblock Darts: Differentiable architecture search.
\newblock In {\em ICLR}, 2019.

\bibitem{seminas}
Renqian Luo, Xu~Tan, Rui Wang, Tao Qin, Enhong Chen, and Tie-Yan Liu.
\newblock Semi-supervised neural architecture search.
\newblock In {\em NeurIPS}, 2020.

\bibitem{dehb}
Neeratyoy Mallik and Noor Awad.
\newblock Dehb: Evolutionary hyperband for scalable, robust and efficient
  hyperparameter optimization.
\newblock In {\em IJCAI}, 2021.

\bibitem{nasbenchasr}
Abhinav Mehrotra, Alberto Gil C.~P. Ramos, Sourav Bhattacharya, {\L}ukasz
  Dudziak, Ravichander Vipperla, Thomas Chau, Mohamed~S Abdelfattah, Samin
  Ishtiaq, and Nicholas~Donald Lane.
\newblock Nas-bench-asr: Reproducible neural architecture search for speech
  recognition.
\newblock In {\em ICLR}, 2021.

\bibitem{penntreebank}
Tom{\'a}{\v{s}} Mikolov, Martin Karafi{\'a}t, Luk{\'a}{\v{s}} Burget, Jan
  {\v{C}}ernock{\`y}, and Sanjeev Khudanpur.
\newblock Recurrent neural network based language model.
\newblock In {\em Annual conference of the international speech communication
  association}, 2010.

\bibitem{miller1989designing}
Geoffrey~F Miller, Peter~M Todd, and Shailesh~U Hegde.
\newblock Designing neural networks using genetic algorithms.
\newblock In {\em ICGA}, volume~89, pages 379--384, 1989.

\bibitem{negrinho2017deeparchitect}
Renato Negrinho and Geoff Gordon.
\newblock Deeparchitect: Automatically designing and training deep
  architectures.
\newblock {\em arXiv preprint arXiv:1704.08792}, 2017.

\bibitem{vu2020bayesian}
Vu~Nguyen, Sebastian Schulze, and Michael Osborne.
\newblock Bayesian optimization for iterative learning.
\newblock In {\em NeurIPS}, 2020.

\bibitem{ning2020surgery}
Xuefei Ning, Wenshuo Li, Zixuan Zhou, Tianchen Zhao, Yin Zheng, Shuang Liang,
  Huazhong Yang, and Yu~Wang.
\newblock A surgery of the neural architecture evaluators.
\newblock {\em arXiv preprint arXiv:2008.03064}, 2020.

\bibitem{oh2019combinatorial}
Changyong Oh, Jakub~M Tomczak, Efstratios Gavves, and Max Welling.
\newblock Combinatorial bayesian optimization using the graph cartesian
  product.
\newblock {\em arXiv preprint arXiv:1902.00448}, 2019.

\bibitem{ottelander2020local}
T~Den Ottelander, Arkadiy Dushatskiy, Marco Virgolin, and Peter~AN Bosman.
\newblock Local search is a remarkably strong baseline for neural architecture
  search.
\newblock In {\em International Conference on Evolutionary Multi-Criterion
  Optimization}, 2021.

\bibitem{paszke2019pytorch}
Adam Paszke, Sam Gross, Francisco Massa, Adam Lerer, James Bradbury, Gregory
  Chanan, Trevor Killeen, Zeming Lin, Natalia Gimelshein, Luca Antiga, et~al.
\newblock Pytorch: An imperative style, high-performance deep learning library.
\newblock In {\em Proceedings of the Annual Conference on Neural Information
  Processing Systems (NeurIPS)}, 2019.

\bibitem{patterson2021carbon}
David Patterson, Joseph Gonzalez, Quoc Le, Chen Liang, Lluis-Miquel Munguia,
  Daniel Rothchild, David So, Maud Texier, and Jeff Dean.
\newblock Carbon emissions and large neural network training.
\newblock {\em arXiv preprint arXiv:2104.10350}, 2021.

\bibitem{peng2020cream}
Houwen Peng, Hao Du, Hongyuan Yu, Qi~Li, Jing Liao, and Jianlong Fu.
\newblock Cream of the crop: Distilling prioritized paths for one-shot neural
  architecture search.
\newblock In {\em NeurIPS}, 2020.

\bibitem{enas}
Hieu Pham, Melody~Y Guan, Barret Zoph, Quoc~V Le, and Jeff Dean.
\newblock Efficient neural architecture search via parameter sharing.
\newblock In {\em ICML}, 2018.

\bibitem{real2019regularized}
Esteban Real, Alok Aggarwal, Yanping Huang, and Quoc~V Le.
\newblock Regularized evolution for image classifier architecture search.
\newblock In {\em AAAI}, 2019.

\bibitem{ru2020revisiting}
Binxin Ru, Clare Lyle, Lisa Schut, Mark van~der Wilk, and Yarin Gal.
\newblock Revisiting the train loss: an efficient performance estimator for
  neural architecture search.
\newblock {\em arXiv preprint arXiv:2006.04492}, 2020.

\bibitem{nasbowl}
Binxin Ru, Xingchen Wan, Xiaowen Dong, and Michael Osborne.
\newblock Neural architecture search using bayesian optimisation with
  weisfeiler-lehman kernel.
\newblock In {\em ICLR}, 2021.

\bibitem{ruchte2020naslib}
Michael Ruchte, Arber Zela, Julien Siems, Josif Grabocka, and Frank Hutter.
\newblock Naslib: a modular and flexible neural architecture search library,
  2020.

\bibitem{sciuto2019evaluating}
Christian Sciuto, Kaicheng Yu, Martin Jaggi, Claudiu Musat, and Mathieu
  Salzmann.
\newblock Evaluating the search phase of neural architecture search.
\newblock In {\em ICLR}, 2020.

\bibitem{scott2015multivariate}
David~W Scott.
\newblock {\em Multivariate density estimation: theory, practice, and
  visualization}.
\newblock John Wiley \& Sons, 2015.

\bibitem{shi2020bonas}
Han Shi, Renjie Pi, Hang Xu, Zhenguo Li, James Kwok, and Tong Zhang.
\newblock Bridging the gap between sample-based and one-shot neural
  architecture search with bonas.
\newblock In {\em NeurIPS}, 2020.

\bibitem{nasbench301}
Julien Siems, Lucas Zimmer, Arber Zela, Jovita Lukasik, Margret Keuper, and
  Frank Hutter.
\newblock Nas-bench-301 and the case for surrogate benchmarks for neural
  architecture search.
\newblock {\em arXiv preprint arXiv:2008.09777}, 2020.

\bibitem{nb301response}
Julien Siems, Lucas Zimmer, Arber Zela, Jovita Lukasik, Margret Keuper, and
  Frank Hutter.
\newblock Nas-bench-301 and the case for surrogate benchmarks for neural
  architecture search: Openreview response, 2021.

\bibitem{stanley2002evolving}
Kenneth~O Stanley and Risto Miikkulainen.
\newblock Evolving neural networks through augmenting topologies.
\newblock {\em Evolutionary computation}, 10(2):99--127, 2002.

\bibitem{swersky2014freeze}
Kevin Swersky, Jasper Snoek, and Ryan~Prescott Adams.
\newblock Freeze-thaw bayesian optimization.
\newblock {\em arXiv preprint arXiv:1406.3896}, 2014.

\bibitem{npenas}
Chen Wei, Chuang Niu, Yiping Tang, and Jimin Liang.
\newblock Npenas: Neural predictor guided evolution for neural architecture
  search.
\newblock {\em arXiv preprint arXiv:2003.12857}, 2020.

\bibitem{wen2019neural}
Wei Wen, Hanxiao Liu, Hai Li, Yiran Chen, Gabriel Bender, and Pieter-Jan
  Kindermans.
\newblock Neural predictor for neural architecture search.
\newblock In {\em ECCV}, 2020.

\bibitem{white2020study}
Colin White, Willie Neiswanger, Sam Nolen, and Yash Savani.
\newblock A study on encodings for neural architecture search.
\newblock In {\em NeurIPS}, 2020.

\bibitem{bananas}
Colin White, Willie Neiswanger, and Yash Savani.
\newblock Bananas: Bayesian optimization with neural architectures for neural
  architecture search.
\newblock In {\em AAAI}, 2021.

\bibitem{white2020local}
Colin White, Sam Nolen, and Yash Savani.
\newblock Local search is state of the art for nas benchmarks.
\newblock In {\em UAI}, 2021.

\bibitem{white2021powerful}
Colin White, Arber Zela, Binxin Ru, Yang Liu, and Frank Hutter.
\newblock How powerful are performance predictors in neural architecture
  search?
\newblock {\em arXiv preprint arXiv:2104.01177}, 2021.

\bibitem{wright1921correlation}
Sewall Wright.
\newblock Correlation and causation.
\newblock {\em Journal of Agricultural Research}, 20:557--580, 1921.

\bibitem{xie2020weight}
Lingxi Xie, Xin Chen, Kaifeng Bi, Longhui Wei, Yuhui Xu, Zhengsu Chen, Lanfei
  Wang, An~Xiao, Jianlong Chang, Xiaopeng Zhang, et~al.
\newblock Weight-sharing neural architecture search: A battle to shrink the
  optimization gap.
\newblock {\em arXiv preprint arXiv:2008.01475}, 2020.

\bibitem{pcdarts}
Yuhui Xu, Lingxi Xie, Xiaopeng Zhang, Xin Chen, Guo-Jun Qi, Qi~Tian, and
  Hongkai Xiong.
\newblock Pc-darts: Partial channel connections for memory-efficient
  architecture search.
\newblock In {\em ICLR}, 2019.

\bibitem{yan2021cate}
Shen Yan, Kaiqiang Song, Fei Liu, and Mi~Zhang.
\newblock Cate: Computation-aware neural architecture encoding with
  transformers.
\newblock In {\em ICML}, 2021.

\bibitem{yan2020does}
Shen Yan, Yu~Zheng, Wei Ao, Xiao Zeng, and Mi~Zhang.
\newblock Does unsupervised architecture representation learning help neural
  architecture search?
\newblock In {\em NeurIPS}, 2020.

\bibitem{yang2019evaluation}
Antoine Yang, Pedro~M Esperan{\c{c}}a, and Fabio~M Carlucci.
\newblock Nas evaluation is frustratingly hard.
\newblock In {\em ICLR}, 2020.

\bibitem{nasbench}
Chris Ying, Aaron Klein, Esteban Real, Eric Christiansen, Kevin Murphy, and
  Frank Hutter.
\newblock Nas-bench-101: Towards reproducible neural architecture search.
\newblock In {\em ICML}, 2019.

\bibitem{you2020greedynas}
Shan You, Tao Huang, Mingmin Yang, Fei Wang, Chen Qian, and Changshui Zhang.
\newblock Greedynas: Towards fast one-shot nas with greedy supernet.
\newblock In {\em CVPR}, 2020.

\bibitem{yu2021landmark}
Kaicheng Yu, Rene Ranftl, and Mathieu Salzmann.
\newblock Landmark regularization: Ranking guided super-net training in neural
  architecture search.
\newblock In {\em CVPR}, 2021.

\bibitem{zela2020understanding}
Arber Zela, Thomas Elsken, Tonmoy Saikia, Yassine Marrakchi, Thomas Brox, and
  Frank Hutter.
\newblock Understanding and robustifying differentiable architecture search.
\newblock In {\em ICLR}, 2020.

\bibitem{zela2020bench}
Arber Zela, Julien Siems, and Frank Hutter.
\newblock Nas-bench-1shot1: Benchmarking and dissecting one-shot neural
  architecture search.
\newblock In {\em ICLR}, 2020.

\bibitem{zhang2020deeper}
Yuge Zhang, Zejun Lin, Junyang Jiang, Quanlu Zhang, Yujing Wang, Hui Xue, Chen
  Zhang, and Yaming Yang.
\newblock Deeper insights into weight sharing in neural architecture search.
\newblock {\em arXiv preprint arXiv:2001.01431}, 2020.

\bibitem{zoph2017neural}
Barret Zoph and Quoc~V. Le.
\newblock Neural architecture search with reinforcement learning.
\newblock In {\em ICLR}, 2017.

\bibitem{zoph2018learning}
Barret Zoph, Vijay Vasudevan, Jonathon Shlens, and Quoc~V Le.
\newblock Learning transferable architectures for scalable image recognition.
\newblock In {\em CVPR}, 2018.

\end{thebibliography}
\bibliographystyle{plain}


\newpage
\appendix

\section{NAS Best Practices Checklist}\label{app:nas_checklist}

In the past few years, the NAS community has called for improving the reproducibility and 
fairness in experimental comparisons~\citep{randomnas, nasbench, yang2019evaluation}.
Recently, a NAS best practices checklist was released~\citep{lindauer2019best}.
We answer each question from this checklist below.


\begin{enumerate}

\item \textbf{Best Practices for Releasing Code}\\[0.2cm]
For all experiments you report: 
\begin{enumerate}
  \item Did you release code for the training pipeline used to evaluate the final architectures?
    \answerNA{We used the training pipelines from NAS-Bench-101, NAS-Bench- 301, 
    and NAS-Bench-NLP, and the code for all three are already publicly available.}
  \item Did you release code for the search space
  \answerNA{We used the search spaces from NAS-Bench-101, NAS-Bench- 301, 
    and NAS-Bench-NLP, and the code for all three are already publicly available.}
  \item Did you release the hyperparameters used for the final evaluation pipeline, as well as random seeds?
  \answerNA{As with prior work that use NAS-Bench-101, NAS-Bench- 301, and NAS-Bench-NLP, 
  our final evaluation pipeline is identical to the training pipeline. 
  Since we averaged over 30 trials of each experiment, we did not report random seeds.}
  \item Did you release code for your NAS method?
  \answerYes{Our code is available at \url{https://github.com/automl/nas-bench-x11}.}
  \item Did you release hyperparameters for your NAS method, as well as random seeds?
  \answerYes{Our code includes runner files with the same hyperparameters and seeds from our paper.}   
\end{enumerate}

\item \textbf{Best practices for comparing NAS methods}
\begin{enumerate}
  \item For all NAS methods you compare, did you use exactly the same NAS benchmark, including the same dataset (with the same training-test split), search space and code for training the architectures and hyperparameters for that code?
    \answerYes{This is true automatically because we only used NAS benchmarks, which fix the training and
    evaluation protocols.}
  \item Did you control for confounding factors (different hardware, versions of DL libraries, different runtimes for the different methods)?
    \answerYes{This is true automatically because we only used NAS Benchmarks, which fix the training and
    evaluation protocols.}	
    \item Did you run ablation studies?
    \answerYes{In Sections~\ref{sec:lce} and \ref{subsec:ablation}, we run ablation studies for the
    our LCE framework. In Section \ref{app:surrogate}, we run ablation studies for our 
    surrogate benchmarks.}
	\item Did you use the same evaluation protocol for the methods being compared?
    \answerYes{This is true automatically because we only used NAS Benchmarks, which fix the training and
    evaluation protocols.}	
	\item Did you compare performance over time?
    \answerYes{We did compare performance over time.}
	\item Did you compare to random search?
    \answerYes{We did compare to random search.}
	\item Did you perform multiple runs of your experiments and report seeds?
    \answerYes{We ran 30 trials for each experiment.}
	\item Did you use tabular or surrogate benchmarks for in-depth evaluations?
    \answerYes{We did use NAS benchmarks.}

\end{enumerate}

\item \textbf{Best practices for reporting important details}
\begin{enumerate}
  \item Did you report how you tuned hyperparameters, and what time and resources
this required?
    \answerYes{We did report information on tuning hyperparameters, including an ablation study
    in Section~\ref{subsec:ablation}.}
  \item Did you report the time for the entire end-to-end NAS method
(rather than, e.g., only for the search phase)?
    \answerYes{Our results include the end-to-end NAS time.}
  \item Did you report all the details of your experimental setup?
    \answerYes{We include all details of our experimental setup.}

\end{enumerate}

\end{enumerate}

\section{Related Work Continued} \label{app:relatedwork}

In this section, we give a more detailed discussion of related work 
(a superset of the related work discussed in Section~\ref{sec:relatedwork}).

NAS has been studied since at least the late
1980s~\citep{miller1989designing, kitano1990designing, stanley2002evolving} 
and has recently seen a 
resurgence~\citep{zoph2017neural,negrinho2017deeparchitect,enas,nasbot,real2019regularized,hu2019forwardnas}.
Early techniques included reinforcement learning~\citep{zoph2017neural, enas},
regularized evolution~\citep{real2019regularized}, and Bayesian optimization~\citep{nasbot}.
Recently, weight sharing~\citep{enas, darts} has become a popular approach to substantially
speed up the runtime of NAS. In this approach, an over-parameterized supernetwork is trained, which
can represent all architectures in the search space. Then all architectures in the search
space can be evaluated using the shared weights.
Some work has claimed that the shared weights are sometimes not effective at ranking 
architectures~\citep{sciuto2019evaluating, zela2020bench, zhang2020deeper}, 
however, weight sharing techniques still achieve strong overall NAS 
performance~\citep{zela2020understanding, li2020geometry}.

Recently, many works have been devoted to performance
prediction~\citep{wen2019neural, ning2020surgery, shi2020bonas, yan2020does, seminas, white2021powerful, nasbowl, bananas} and multi-fidelity techniques~\citep{bohb} which has reduced
the runtime gap between iterative and weight sharing techniques. 
For detailed surveys on NAS, see~\citep{nas-survey,xie2020weight}. 
The most widely used type of search space in prior work is the cell-based search space~\citep{zoph2018learning,liu2017progressive}, where the architecture search is over a relatively small directed acyclic graph representing an architecture.

\paragraph{Learning curve extrapolation.}
Several methods have been proposed to estimate the final validation accuracy
of a neural network by extrapolating the learning curve of a partially trained
neural network.
Techniques include, fitting the partial curve to an ensemble of parametric
functions~\citep{domhan2015speeding}, predicting the performance based on the partial trained neural network configurations~\citep{baker2017accelerating}, summing the training losses~\citep{ru2020revisiting},
using the basis functions as the output layer of a Bayesian neural network~\citep{lcnet}, using previous learning curves as basis function extrapolators~\citep{chandrashekaran2017speeding}, using the positive-definite covariance kernel to capture a variety of training curves~\citep{swersky2014freeze},  or using a Bayesian recurrent neural network~\citep{gargiani2019probabilistic}. While in this work we focus on multi-fidelity optimization utilizing learning curve-based extrapolation, another main category of methods lie in bandit-based algorithm selection~\citep{hyperband,bohb,abohb,huang2020asymptotically,dehb}, and the fidelities can be further adjusted according to the previous observations or a learning rate scheduler~\citep{k2016multifidelity,k2017multifidelity,klein17fast}.

\paragraph{NAS benchmarks.}
NAS-Bench-101~\citep{nasbench}, a tabular NAS benchmark, was created by defining a 
search space of size $423\,624$ unique architectures and then training all architectures from the search space on CIFAR-10 until 108 epochs. However, the train, validation, and test accuracies
are only reported for epochs 4, 12, 36, and 108, and the train/valid/test losses are not reported. NAS-Bench-1shot1~\citep{zela2020bench} defines a subset of 
the NAS-Bench-101 search space that allows one-shot algorithms to be run.
NAS-Bench-201~\citep{nasbench201} contains $15\,625$
architectures, of which $6\,466$ are unique up to isomorphisms.
It comes with full learning curve information on three datasets:
CIFAR-10~\citep{CIFAR10}, CIFAR-100~\citep{CIFAR10},
and ImageNet16-120~\citep{tinyimagenet17}.
Recently, NAS-Bench-201 was extended to NATS-Bench~\citep{natsbench} which
searches over architecture size as well as architecture topology.

Virtually every published NAS method for image classification in the last 3 years evaluates on the DARTS search space with CIFAR-10~\citep{nb301response}.
The DARTS search space ~\citep{darts} consists of $10^{18}$ neural architectures,
making it computationally prohibitive to create a tabular benchmark. 
To overcome this fundamental limitation and query architectures in this much larger search space, 
NAS-Bench-301~\citep{nasbench301} evaluates various regression models 
trained on a sample of $60\,000$ architectures that is carefully created to cover the whole search 
space. The surrogate models allow users to query the validation accuracy (at epoch 100)
and training time for any of the $10^{18}$ architectures in the DARTS search space.
However, since the surrogates do not predict the entire learning curve, it is not possible to run
multi-fidelity algorithms.

NAS-Bench-NLP~\citep{nasbenchnlp} is a search space for language modeling tasks.
The search space consists of $10^{53}$ LSTM-like architectures, of which
$14\,322$ are evaluated on Penn Tree Bank~\citep{penntreebank}, containing the training, validation,
and test losses/accuracies from epochs 1 to 50. Since only $14\,322$ of $10^{53}$
architectures can be queried, this dataset cannot be directly used for NAS experiments.
NAS-Bench-ASR~\citep{nasbenchasr} is a recent tabular NAS benchmark for speech recognition.
The search space consists of $8\,242$ architectures with full learning curve information.
For an overview of NAS benchmarks, see Table~\ref{tab:benchmarks}.
We give a more detailed discussion of all related work in Appendix~\ref{app:relatedwork}.

\section{Details from Section~\ref{sec:surrogate}} \label{app:surrogate}

In this section, we give more details from Section~\ref{sec:surrogate}, 
and we present a full ablation study.

Recall the following notation, repeated from Section~\ref{sec:surrogate}.
Given a search space $\D$, let $(\vec{x}_i, \vec{y}_i) \sim \D$ denote one datapoint, 
where $\vec{x}_i \in \R^d$ is the architecture encoding, 
and $\vec{y}_i \in [0, 1]^{\Emax}$  is a learning curve of validation accuracies drawn 
from a distribution $Y(\vec x_i)$ based on training the architecture for $\Emax$ epochs on 
a fixed training pipeline with a random initial seed.
For each learning curve $\vec{y}_i$, we have 
$\vec{y}_i = \E[Y(\vec x_i)] + \vec{\epsilon}_i$,
where $\E[Y(\vec x_i)]\in [0, 1]^{\Emax}$ is fixed and depends only on $\vec{x}_i$,
and $\vec{\epsilon}_i\in [0, 1]^{\Emax}$ comes from a noise distribution
$Z_i$ with expectation 0 for all epochs.
In practice, $\E[Y(\vec x_i)]$ can be estimated by averaging a large set of learning curves 
produced by training architecture $\vec{x}_i$ with different initial seeds. 
We represent such an estimate as $\bar{\vec y}_i$. 
As explained in Section~\ref{sec:surrogate},
we split the surrogate model creation into two parts:
we train a model $f: \R^d \to [0,1]^{\Emax}$ to predict the deterministic part of the learning curve, $\bar{\vec y}_i$, 
and we train a noise model $p_\phi (\vec \epsilon \mid \bar{\vec y}, \vec x)$,
parameterized by $\phi$, to simulate the random draws from $Z_i$. 
See Figure~\ref{fig:summary} for a summary of our entire surrogate creation method
(assuming SVD).

In Section~\ref{sec:surrogate}, we described singular value decomposition (SVD)
as a technique to create the compression and decompression functions
$c_k:[0, 1]^{\Emax}\rightarrow [0, 1]^k$ and $d_k :[0, 1]^k\rightarrow [0, 1]^{\Emax}$, 
respectively, for $k \ll \Emax$, which aid in the creation of a model $f$.
Now we describe our second technique for compression and decompression:
a variational autoencoder (VAE)~\citep{kingma2013auto}.
The VAE has the benefit over SVD that it is a non-linear dimensionality reduction technique.
However, it is harder to train as it has more hyperparameters, as opposed to SVD which only
has $k$ as a parameter. We fit a VAE model to the learning curves.
We used a simple PyTorch~\citep{paszke2019pytorch} implementation of VAE
which has four fully connected layers of 512 nodes, separated by ReLU, in both the
encoder and decoder architecture. Finally, we add dropout of 0.2 and the Adam optimizer.
We constrain the dimension of the bottleneck latent space the same amount as with SVD:
$k=5$. This creates a non-linear dimensionality reduction model.
In Table~\ref{tab:surrogate_ablation}, we see that the VAE does not perform as well as SVD.

As mentioned in Section~\ref{sec:surrogate}, we try three different models for the
main $\mu$-model (which predicts the compressed learning curves from the architecture
encodings): LGBoost~\citep{ke2017lightgbm}, XGBoost~\citep{chen2016xgboost}, and a standard
multilayer perceptron (MLP). For LGBoost and XGBoost, we use the default parameters reported
from NAS-Bench-301~\citep{nasbench301}. For MLP, we use one layer with 64 nodes, with SGD with
learning rate 0.001. We also tried five layers, which performed worse.

\begin{figure}[t]
	\centering
	\includegraphics[width=0.32\textwidth]{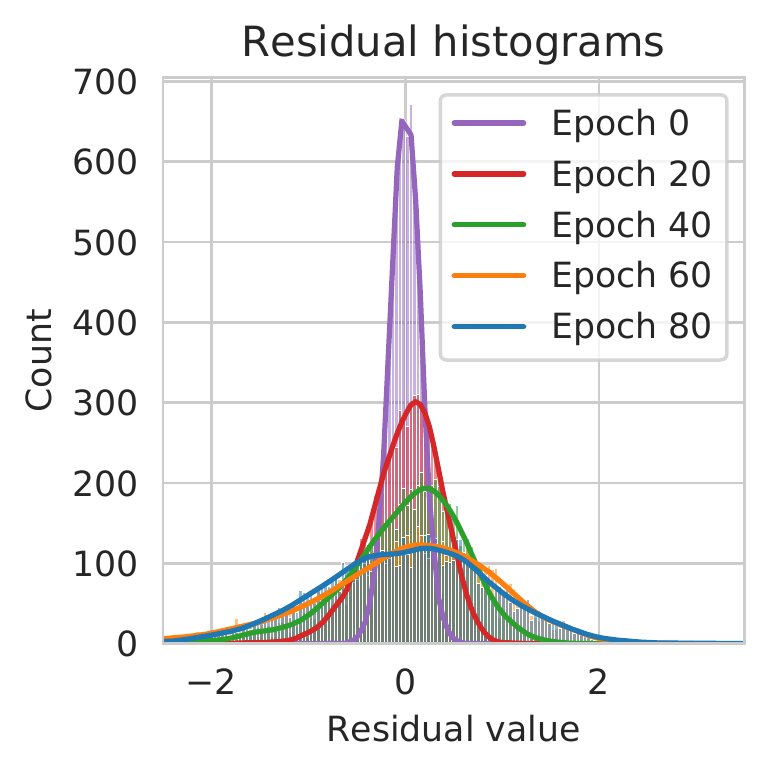}
	\includegraphics[width=0.32\textwidth]{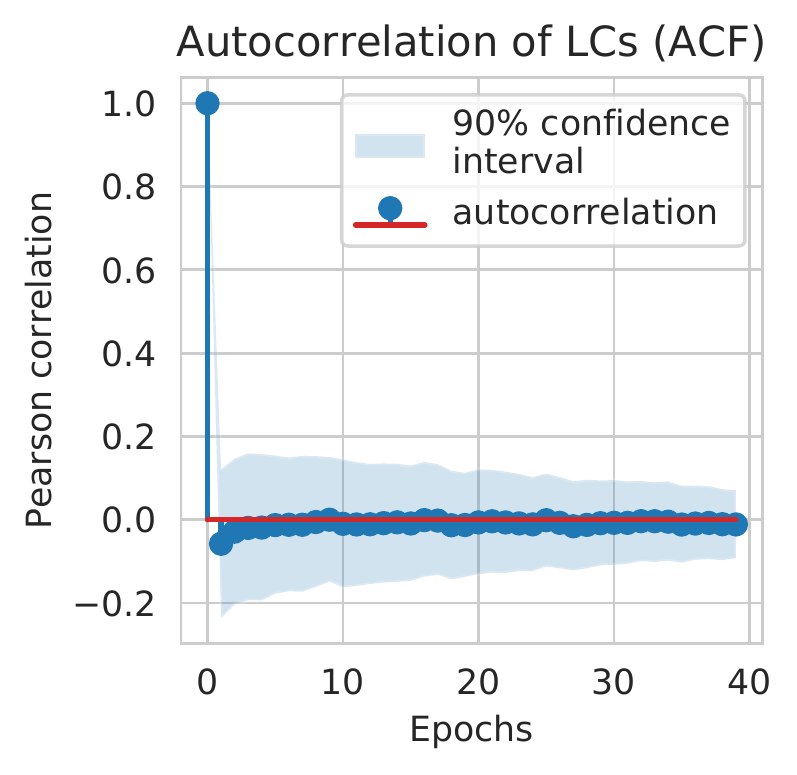}
	\includegraphics[width=0.32\textwidth]{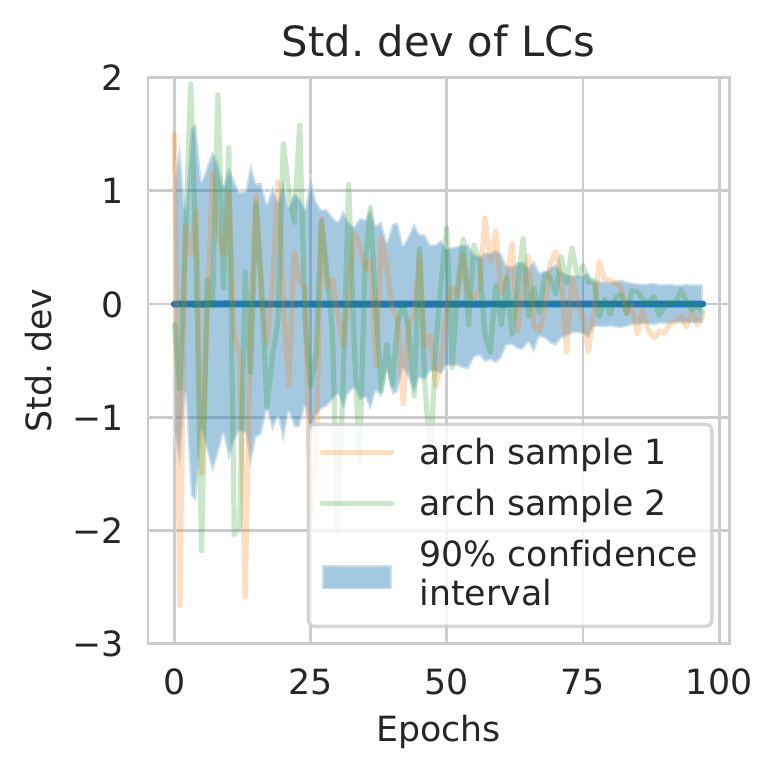}
	\caption{
	A plot of the residuals across all architectures for five different epochs (left).
	We see that the distributions are roughly Gaussian.
	A plot of the autocorrelation function (ACF) averaged over all training learning curves
	(middle). We see that there is only a small amount of autocorrelation.
	A plot of the 90\% confidence intervals of the residuals at each epoch (right).
	All plots use the NAS-Bench-301 learning curve training set.
	}
	\label{fig:noise}
\end{figure}

Finally, we consider three different noise models as described in Section~\ref{sec:surrogate}.
Recall that we create a new dataset of predicted $\vec{\epsilon}_i$ values, which we call residuals,
by subtracting the reconstructed mean learning curves from the real learning curves in
$\Dtr$. That is, $\hat {\vec \epsilon}_i = \vec y_i - (d_k \circ c_k) \left(\vec{y}_i\right)$ is the 
residual for the $i$th learning curve. Our three noise models are based on two different assumptions:
\emph{(1)} the noise distribution is the same for all architectures, 
and \emph{(2)} for each architecture, the noise in a small window of epochs are iid.
Now we evaluate these assumptions. In Figure~\ref{fig:noise} (left), we plot the residuals
from the NAS-Bench-301 training set at five different epochs, 
showing that the distributions are roughly Gaussian, across all architectures.
Recall that noise model \emph{(i)} is a simple standard deviation statistic computed for
each epoch independently, across all architectures. In Figure~\ref{fig:noise} (middle),
we plot the autocorrelation function (ACF) averaged over all training learning curves
on NAS-Bench-301. We see that there is very little autocorrelation in the learning curves, 
which justifies the use of the first noise model. Recall that our second noise model
uses Gaussian kernel density estimation (GKDE)~\citep{scott2015multivariate}
across all learning curves. This is essentially the same as the first noise model but with
the ability to capture the small amount of autocorrelation present.
Finally, recall that our our third noise model does not assume that the residual distribution
is similar across all architectures. Instead, it estimates the standard deviation for
each epoch using a sliding window of size 10 across the epochs for each architecture.
See Figure~\ref{fig:noise} (right) for the 90\% confidence intervals of the residuals at
each epoch on NAS-Bench-301. Although the sliding window noise model has the benefit of capturing 
different distributions for different architectures, we see that the standard deviation
steadily decreases as the epoch number increases, meaning that the noise in a small window of
epochs are not perfectly iid.

In Table~\ref{tab:surrogate_ablation}, we run a full ablation study by testing all eighteen
combinations of \{SVD, VAE\}, \{LGB, XGB, MLP\}, and \{GKDE, STD, window\} for NAS-Bench-111
and NAS-Bench-311. Recall that, as explained in Section~\ref{sec:surrogate}, the first
four metrics evaluate only the prediction of the mean learning curves using a held-out
test set, so the noise model has no effect on the first four metrics (average $R^2$, final
$R^2$, average KT, and final KT). For the final two metrics (average KL and final KL),
we use a test set of learning curves consisting of five seeds of architectures,
so that we can estimate the KL divergence between the real learning curve distribution
and the predicted distribution.
Note that none of the NAS-Bench-NLP architectures were trained more than once, so we
are unable to test the noise models for NAS-Bench-NLP11.
Across NAS-Bench-111, NAS-Bench-311, and NAS-Bench-NLP11, we see that SVD-LGB performs
substantially better than all of the other options for the model.

\begin{table} 
\caption{Evaluation of the surrogate benchmarks on test sets,
with all combinations of models.
For NAS-Bench-111 and NAS-Bench-NLP11, we use architecture accuracies
as additional features to improve performance.
As explained in Section~\ref{app:surrogate}, no architectures in the NAS-Bench-NLP
dataset were trained more than once, so we do not compute KL divergence for
NAS-Bench-NLP11.
} \label{tab:surrogate_ablation}
\centering
{\small
\begin{tabular}{@{}l|c|c|c|c|c|c@{}}
\toprule
\multicolumn{1}{l}{\textbf{Benchmark}} & \multicolumn{1}{c}{\textbf{Avg.\ $R^2$}} & \multicolumn{1}{c}{\textbf{Final $R^2$}} & \multicolumn{1}{c}{\textbf{Avg.\ KT}} & \multicolumn{1}{c}{\textbf{Final KT}} & \multicolumn{1}{c}{\textbf{Avg.\ KL}} & \multicolumn{1}{c}{\textbf{Final KL}}\\
\midrule 
NAS-Bench-111 \\
\midrule
SVD-LGB-GKDE & \textbf{0.630} & \textbf{0.853} & \textbf{0.611} & \textbf{0.794} & \textbf{1.641} & 0.516 \\
SVD-LGB-STD & \textbf{0.630} & \textbf{0.853} & \textbf{0.611} & \textbf{0.794} & 2.768 & \textbf{0.383} \\
SVD-LGB-window & \textbf{0.630} & \textbf{0.853} & \textbf{0.611} & \textbf{0.794} & 24.402 & 3.303 \\
SVD-XGB-GKDE & 0.329 & 0.378 & 0.408 & 0.429 & 2.743 & 0.580 \\
SVD-XGB-STD & 0.329 & 0.378 & 0.408 & 0.429 & 4.867 & 0.503 \\
SVD-XGB-window & 0.329 & 0.378 & 0.408 & 0.429 & 38.457 & 16.172 \\
SVD-MLP-GKDE & 0.195 & 0.065 & 0.330 & 0.290 & 4.599 & 0.762 \\
SVD-MLP-STD & 0.195 & 0.065 & 0.330 & 0.290 & 8.417 & 0.848 \\
SVD-MLP-window & 0.195 & 0.065 & 0.330 & 0.290 & 82.180 & 15.711 \\
VAE-LGB-GKDE & 0.267 & 0.218 & 0.462 & 0.617 & 3.788 & 0.829 \\
VAE-LGB-STD & 0.267 & 0.218 & 0.462 & 0.617 & 6.866 & 0.972 \\
VAE-LGB-window & 0.267 & 0.218 & 0.462 & 0.617 & 53.866 & 19.820 \\
VAE-XGB-GKDE & 0.311 & 0.272 & 0.453 & 0.559 & 3.828 & 0.828 \\
VAE-XGB-STD & 0.311 & 0.272 & 0.453 & 0.559 & 6.940 & 0.969 \\
VAE-XGB-window & 0.311 & 0.272 & 0.453 & 0.559 & 55.654 & 19.614 \\
VAE-MLP-GKDE & 0.218 & 0.007 & 0.386 & 0.369 & 4.583 & 0.844 \\
VAE-MLP-STD & 0.218 & 0.007 & 0.386 & 0.369 & 8.386 & 1.001 \\
VAE-MLP-window & 0.218 & 0.007 & 0.386 & 0.369 & 83.481 & 19.091 \\
\midrule
NAS-Bench-311 \\
\midrule
SVD-LGB-GKDE & \textbf{0.779} & \textbf{0.800} & \textbf{0.728} & \textbf{0.788} & \textbf{0.503} & \textbf{0.548} \\
SVD-LGB-STD & \textbf{0.779} & \textbf{0.800} & \textbf{0.728} & \textbf{0.788} & 0.919 & 1.036 \\
SVD-LGB-window & \textbf{0.779} & \textbf{0.800} & \textbf{0.728} & \textbf{0.788} & 1.566 & 4.083 \\
SVD-XGB-GKDE & 0.522 & 0.546 & 0.607 & 0.654 & 1.783 & 3.272 \\
SVD-XGB-STD & 0.522 & 0.546 & 0.607 & 0.654 & 3.271 & 5.958 \\
SVD-XGB-window & 0.522 & 0.546 & 0.607 & 0.654 & 5.282 & 19.432 \\
SVD-MLP-GKDE & 0.564 & 0.549 & 0.573 & 0.603 & 15.727 & 29.057 \\
SVD-MLP-STD & 0.564 & 0.549 & 0.573 & 0.603 & 28.833 & 52.515 \\
SVD-MLP-window & 0.564 & 0.549 & 0.573 & 0.603 & 45.071 & 167.140 \\
VAE-LGB-GKDE & 0.431 & 0.447 & 0.568 & 0.616 & 5.995 & 13.486 \\
VAE-LGB-STD & 0.431 & 0.447 & 0.568 & 0.616 & 11.015 & 24.836 \\
VAE-LGB-window & 0.431 & 0.447 & 0.568 & 0.616 & 17.510 & 79.773 \\
VAE-XGB-GKDE & 0.397 & 0.427 & 0.577 & 0.624 & 6.520 & 16.739 \\
VAE-XGB-STD & 0.397 & 0.427 & 0.577 & 0.624 & 11.978 & 30.368 \\
VAE-XGB-window & 0.397 & 0.427 & 0.577 & 0.624 & 18.883 & 97.485 \\
VAE-MLP-GKDE & 0.509 & 0.520 & 0.584 & 0.619 & 13.545 & 33.851 \\
VAE-MLP-STD & 0.509 & 0.520 & 0.584 & 0.619 & 24.770 & 61.455 \\
VAE-MLP-window & 0.509 & 0.520 & 0.584 & 0.619 & 38.593 & 196.246 \\
\midrule
NAS-Bench-NLP11 \\
\midrule
SVD-LGB & \textbf{0.906} & \textbf{0.882} & \textbf{0.862} & \textbf{0.820} & - & - \\
SVD-XGB & 0.849 & 0.865 & 0.786 & 0.735 & - & - \\
SVD-MLP & 0.120 & 0.108 & 0.292 & 0.275 & - & - \\
VAE-LGB & 0.789 & 0.795 & 0.802 & 0.747 & - & - \\
VAE-XGB & 0.826 & 0.838 & 0.797 & 0.739 & - & - \\
VAE-MLP & 0.150 & 0.160 & 0.315 & 0.300 & - & - \\
\bottomrule
\end{tabular}
}
\end{table}

\paragraph{Surrogate training details.}
Finally, we give more details for the surrogate training.
For NAS-Bench-111, as discussed in Section~\ref{sec:surrogate}, we created a new set
of trained architectures with the full learning curve information. We kept the training
pipeline nearly the same as in the orginal NAS-Bench-101 repository. However, instead
of the TPU v2 acceleator as in the original work, we used an RTX 3070. We needed to
change the batch size from 256 to 200 to account for this hardware change, which 
we found had a negligible affect on the final accuracy. We trained 1101 new architectures
and used this new set as the new ``ground-truth'' when training and evaluating
NAS-Bench-111. As explained in Section~\ref{sec:surrogate}, the accuracies from the original
NAS-Bench-101 benchmark were used as features to improve the performance of our surrogate,
but not used as ground truth.
In Table \ref{tab:nb111_kts}, we show that the KT values for NAS-Bench-111 are roughly equivalent to those achieved by a 1-seed tabular benchmark.

\begin{table} 
\caption{NAS-Bench-111 rank correlations computed on a separate test set with architectures trained for two different random seeds each. This allows the comparison with the rank correlation of an independent set of ground truth architectures. We find that the NAS-Bench-111 mean model is on par with the ground truth.
} \label{tab:nb111_kts}
\centering
{\small
\begin{tabular}{@{}l|c|c|c|c@{}}
\toprule
\multicolumn{1}{l}{\textbf{Benchmark}} & \multicolumn{1}{c}{\textbf{Avg.\ $R^2$}} & \multicolumn{1}{c}{\textbf{Final $R^2$}} & \multicolumn{1}{c}{\textbf{Avg.\ KT}} & \multicolumn{1}{c}{\textbf{Final KT}} \\
\midrule 
NAS-Bench-111 & 0.557 & 0.541 & \textbf{0.660} & 0.860 \\
Ground truth (1 seed) & \textbf{0.593} & \textbf{0.920} & 0.619 & \textbf{0.873} \\
\bottomrule
\end{tabular}
}
\end{table}

For NAS-Bench-311, training was straightforward. We used the original NAS-Bench-301 dataset,
which already achieves good coverage~\citep{nasbench301}, and we did not use any additional
features.
In Table \ref{tab:nb311_kts}, we show that the mean model in NAS-Bench-311 achieves higher rank correlation even than a set of learning curves averaged over four random seeds, by using a separate test set from the NAS-Bench-301 dataset which evaluates 500 architectures with 5 seeds each.

\begin{table} 
\caption{NAS-Bench-311 rank correlations computed on a separate test set with architectures trained for five different random seeds each. This allows the comparison with sets of learning curves averaged over multiple seeds. We find that the NAS-Bench-311 mean model performs better than a 4-seed mean.
} \label{tab:nb311_kts}
\centering
{\small
\begin{tabular}{@{}l|c|c|c|c@{}}
\toprule
\multicolumn{1}{l}{\textbf{Benchmark}} & \multicolumn{1}{c}{\textbf{Avg.\ $R^2$}} & \multicolumn{1}{c}{\textbf{Final $R^2$}} & \multicolumn{1}{c}{\textbf{Avg.\ KT}} & \multicolumn{1}{c}{\textbf{Final KT}} \\
\midrule 
NAS-Bench-311 & \textbf{0.731} & 0.845 & \textbf{0.637} & \textbf{0.718} \\
Ground truth (1 seed) & 0.534 & 0.782 & 0.508 & 0.641 \\
Ground truth (mean of 2 seeds) & 0.651 & 0.835 & 0.555 & 0.683 \\
Ground truth (mean of 3 seeds) & 0.690 & 0.859 & 0.579 & 0.704 \\
Ground truth (mean of 4 seeds) & 0.710 & \textbf{0.870} & 0.592 & 0.712 \\
\bottomrule
\end{tabular}
}
\end{table}

For NAS-Bench-NLP11, as described earlier, it is challenging to create an accurate surrogate
benchmark because there are only $14\,322$ evaluated architectures for a search space of
total size $10^{53}$. Therefore, we used two techniques to improve performance. First,
we used a subset of the search space, restricting the architectures to a maximum of
12 nodes (reducing the size to $10^{22}$), and we added the validation accuracies from
the first three epochs of training each architecture, as features.
These two techniques were shown to substantially improve the performance of
NAS-Bench-NLP11, as shown in Table~\ref{tab:surrogate}.
On an RTX 3070, training architectures from NAS-Bench-NLP takes about 90 seconds per epoch.
Although adding in the first three epochs substantially improves the accuracy of our
surrogate benchmark, it comes at the cost of query time. While NAS-Bench-111 and NAS-Bench-311
take under one second to query, a query to NAS-Bench-NLP11 now requires training an architecture
for three epochs. Note that this is still a $15\times$ speedup over performing NAS directly 
without a surrogate benchmark.

We also create NAS-Bench-211 to further evaluate our surrogate creation technique (since NAS-Bench-201 already has complete learning curves). We train the surrogate on 90\% of architectures from NAS-Bench-201 (14\,062 architectures) and test on the remaining 10\%. SVD-LGB-window achieves the best performance.
The rank correlation values are on par with a 1-seed tabular benchmark (see Table \ref{tab:nb211_kts}).

\begin{table} 
\caption{NAS-Bench-211 rank correlations computed on a test set with architectures trained for three different random seeds each. This allows the comparison with sets of learning curves averaged over multiple seeds. We find that the NAS-Bench-211 mean model performs on par with 1-seed ground truth.
} \label{tab:nb211_kts}
\centering
{\small
\begin{tabular}{@{}l|c|c|c|c@{}}
\toprule
\multicolumn{1}{l}{\textbf{Benchmark}} & \multicolumn{1}{c}{\textbf{Avg.\ $R^2$}} & \multicolumn{1}{c}{\textbf{Final $R^2$}} & \multicolumn{1}{c}{\textbf{Avg.\ KT}} & \multicolumn{1}{c}{\textbf{Final KT}} \\
\midrule 
NAS-Bench-211 & 0.893 & 0.958 & \textbf{0.701} & 0.842 \\
Ground truth (1 seed) & 0.866 & 0.999 & 0.646 & 0.916 \\
Ground truth (mean of 2 seeds) & \textbf{0.900} & \textbf{0.999} & 0.679 & \textbf{0.926} \\
\bottomrule
\end{tabular}
}
\end{table}

\section{Details from Section~\ref{sec:experiments} (Experiments)}\label{app:experiments}

In this section, we give more details from Section~\ref{sec:experiments}, 
and we present more experiments.

Our work uses existing NAS Benchmarks. In Table~\ref{tab:licenses}, we report the licenses for each one.

\begin{table}
\caption{Licenses for the datasets that we use.}
\centering
\begin{tabular}{@{}l|c|c@{}}
\toprule
\multicolumn{1}{l}{\textbf{Dataset}} & \multicolumn{1}{c}{\textbf{License}} & \multicolumn{1}{c}{\textbf{URL}} \\
\midrule 
NAS-Bench-101 & Apache 2.0 & \url{https://github.com/google-research/nasbench} \\
NAS-Bench-201 & MIT & \url{https://github.com/D-X-Y/NAS-Bench-201} \\
NAS-Bench-301 & Apache 2.0 & \url{https://github.com/automl/nasbench301} \\
NAS-Bench-NLP & None & \url{https://github.com/fmsnew/nas-bench-nlp-release} \\
\bottomrule
\end{tabular}
\label{tab:licenses}
\end{table}

\subsection{LCE Results}
\begin{figure}
\centering
\includegraphics[width=.32\columnwidth]{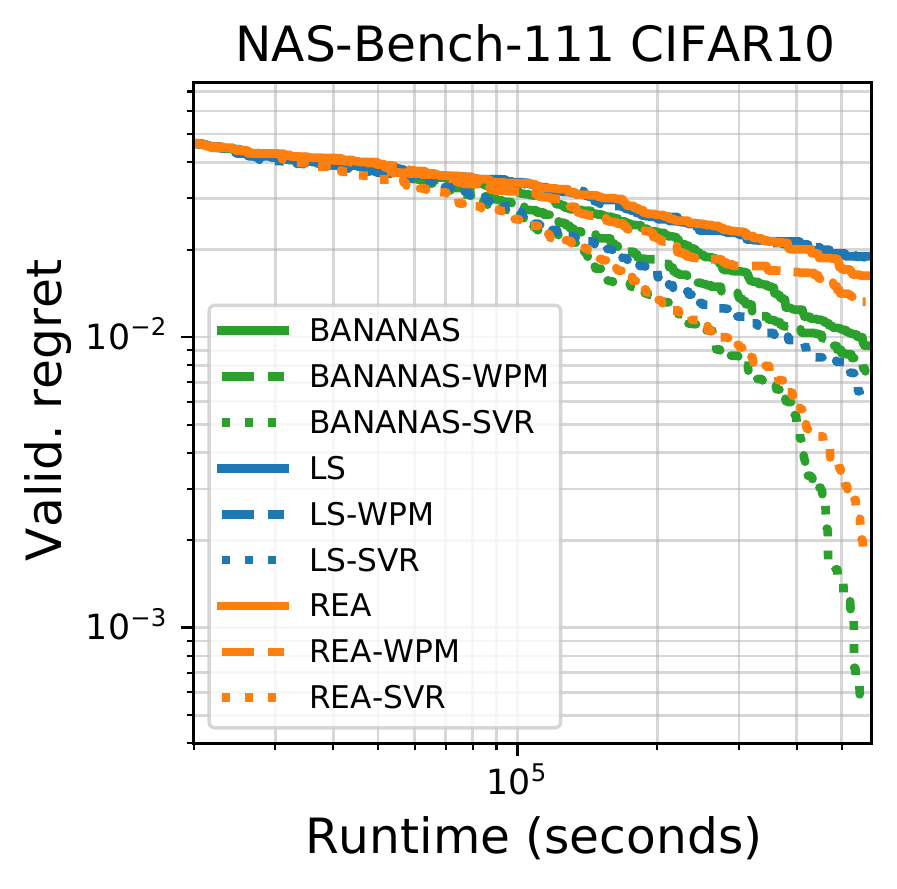}
\includegraphics[width=.32\columnwidth]{fig/lce/nas311_c10_lce_nov2.pdf}
\includegraphics[width=.32\columnwidth]{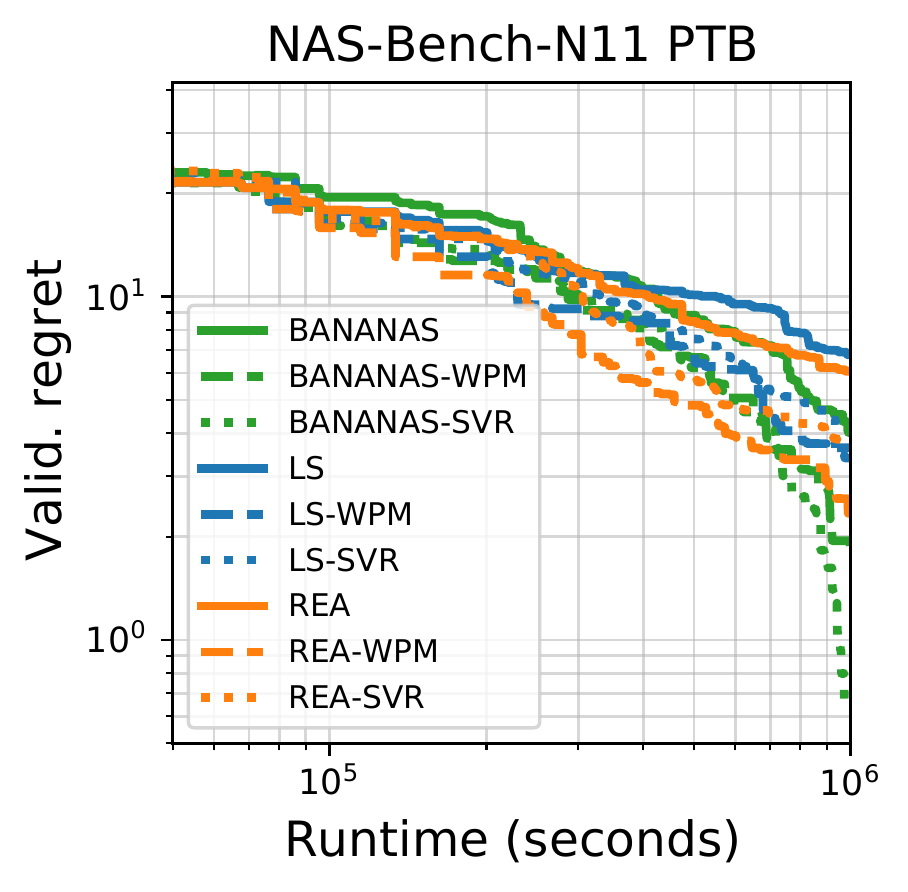}
\includegraphics[width=.32\columnwidth]{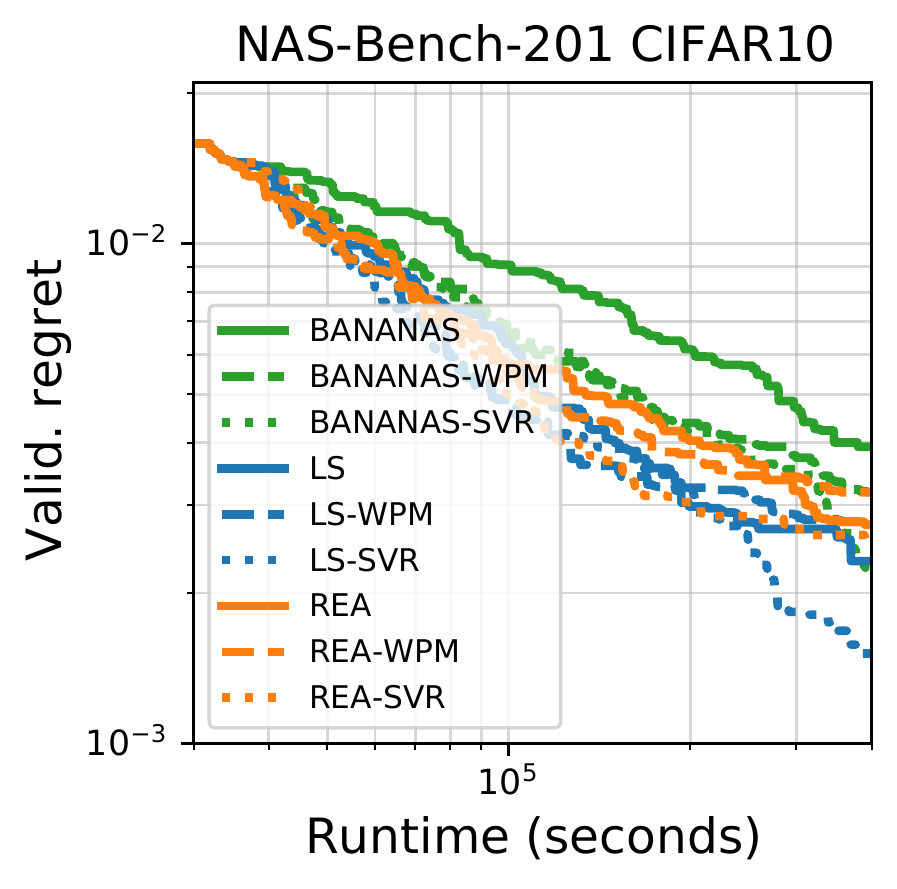}
\includegraphics[width=.32\columnwidth]{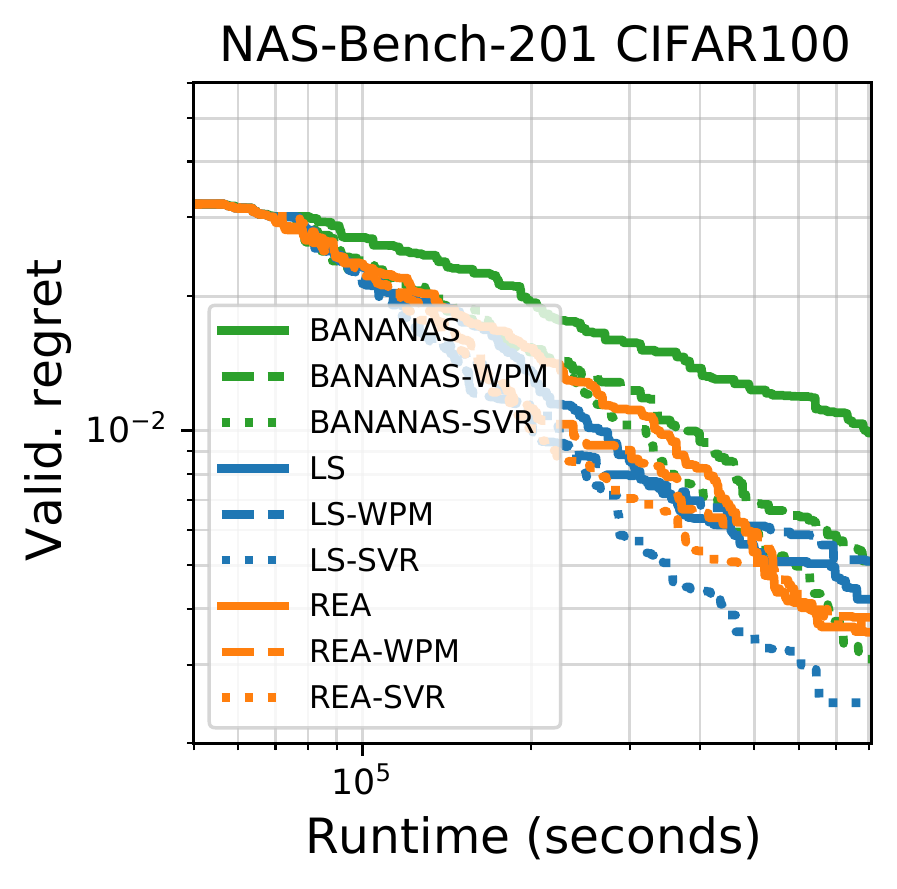}
\includegraphics[width=.32\columnwidth]{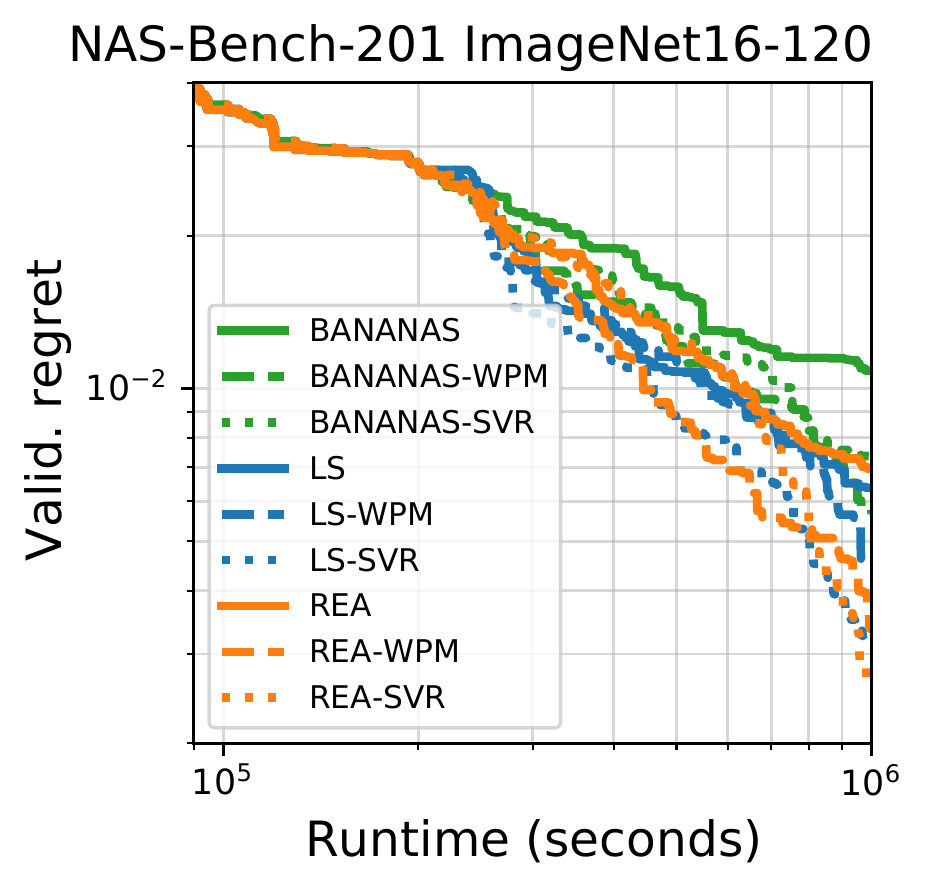}
\caption{LCE Framework applied to single-fidelity algorithms on NAS-Bench-111, NAS-Bench-311,
NAS-Bench-NLP11, and NAS-Bench-201.}
\label{fig:lce_appendix}
\end{figure}

Next, we give the LCE results for four search spaces, which is an extension of the results
from Figure~\ref{fig:lce}.
That is, we test the improvement of three different single-fidelity algorithms
when used with our LCE framework from Section~\ref{sec:lce}, using WPM or SVR as the
LCE techniques.
We see that across all search spaces, for each single-fidelity algorithm, WPM and SVR both
give improvements over the original algorithm, and SVR tends to give the larger improvement 
compared to WPM.
In Figures \ref{fig:nas} and \ref{fig:lce_appendix}, an earlier version of the NAS-Bench-NLP11 noise model was used. We also added slight clipping for NAS-Bench-111 and -311 to reduce the number of spike anomalies as described in Section \ref{subsec:creation}.

\subsection{Ablation study.} \label{subsec:ablation}
\begin{figure}
	\centering
	\includegraphics[width=0.5\textwidth]{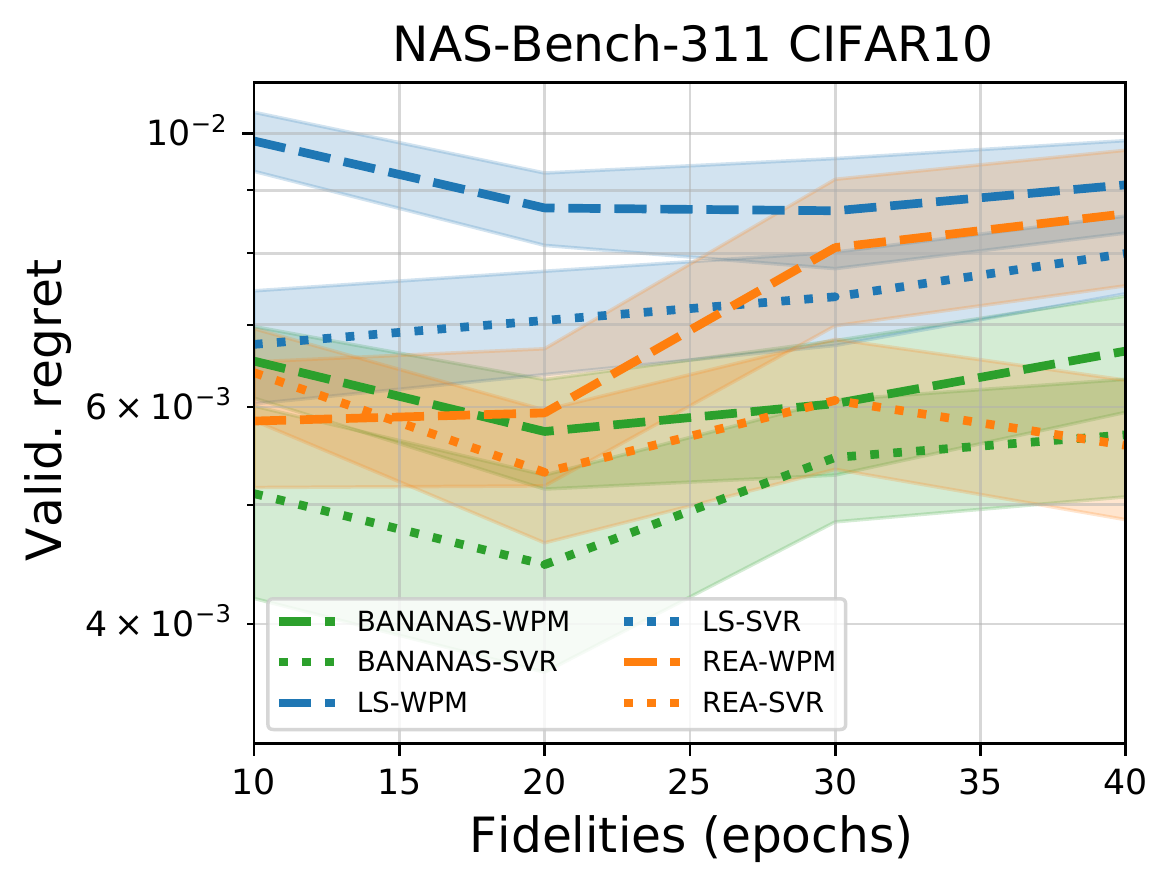}
	\caption{Different fidelities and their effect on NAS performance on NAS-Bench-311. 
	The wall-clock time [s] is set to 2e6. The result are reported across 30 seeds.}
	\label{fig:nas_311_mf_ablation}
\end{figure}

We evaluate the effect of different fidelities on NAS-Bench-311. 
In Figure~\ref{fig:nas_311_mf_ablation},
we plot the validation regret of the SVR and WPM-based algorithms after $2\times 10^6$ seconds,
varying the initial fidelity (epoch) from which the learning curve is extracted, from 10
to 40. That is, the leftmost points run LCE by extrapolating from epoch 10 to epoch 100,
and the rightmost points run LCE by extrapolating from epoch 40 to epoch 100. Note that
there is a tradeoff between time saved (from only evaluating to 10 epochs vs 40) and
accuracy of LCE (extrapolating from 10 epochs is more challenging than from 40 epochs).
We see that overall, epoch 20 performs the best. Notably, BANANAS-SVR and REA-SVR
(two of the best-performing algorithms across all search spaces) achieve top
performance at epoch 20.

\subsection{NAS algorithm descriptions and details}

We give a description and implementation details for each NAS algorithm from 
Section~\ref{sec:experiments}. Note that all algorithms were implemented in
NASLib~\citep{ruchte2020naslib}, keeping the implementation as close as possibe
to the original implementation.

\begin{itemize}
    \item\textbf{Random search.} 
    Random search is a simple baseline which draws architectures at
    random and then returns the architecture with the lowest validation error.
    Note that multiple papers have shown that random search is competitive 
    with other NAS algorithms \cite{randomnas, sciuto2019evaluating}.

    \item\textbf{Local search.} 
    Another baseline, local search has been shown to perform surprisingly 
    well~\citep{white2020local, ottelander2020local}, even on the DARTS
    search space~\citep{nasbench301}. It works by iteratively evaluating all
    architectures in the neighborhood of the current best architecture found so far.
    The neighborhood is defined as the set of architectures which differ by
    one operation or edge.
    We used the implementation from NASLib \citep{ruchte2020naslib}. Notably, this is slightly different from the White et al.~\citep{white2020local} implementation which may explain the worse performance on NAS-Bench-311.

    \item \textbf{BANANAS.} This algorithm~\citep{bananas} 
    is based on Bayesian optimization using an ensemble of three MLPs as the model.
    We use the code directly from the original repository. We set the encoding
    to the adjacency matrix encoding instead of the path encoding. 
    A predictor (trained on all architectures evaluated so far) chooses $k$ 
    architectures which are then evaluated. In our experiments, the candidate pool is created by mutating the top four architectures ten times each (two times for each of the edit distance from one to five),  and we set $k=20$.

    \item\textbf{Regularized evolution.} 
    This algorithm~\citep{real2019regularized} is based on evolution.
    It consists of iteratively mutating the best architectures
    out of a sample of all architectures evaluated so 
    far. A mutation is defined as randomly changing one operation or
    edge. We used the NAS-Bench-101~\citep{nasbench} implementation, changing the
    population size from 50 to 20.

    \item\textbf{Hyperband.} 
    This algorithm~\citep{hyperband} is based on random search with successive
    halving. It is based on successive halving, in which architectures are iteratively 
    trained at a low fidelity, and then only the best-performing architectures are trained 
    for longer in the next iteration, until the maximum number of epochs is reached. 
    Hyperband performs multiple rounds of successive halving at different initial
    fidelities. We use the \texttt{hpbandster} implementation, adapted to
    NASLib~\citep{ruchte2020naslib}.
    
    \item\textbf{Bayesian optimization Hyperband.} 
    This algorithm~\citep{bohb} is based on combining Hyperband with Bayesian optimization.
    It starts the same way as Hyperband, but in the later rounds, for each fidelity
    a KDE model is trained using the trained architectures from previous rounds.
    Then the best architectures are chosen using Bayesian optimization with the model.
    We use the \texttt{hpbandster} implementation, adapted to
    NASLib~\citep{ruchte2020naslib}.

\end{itemize}

\end{document}